\documentclass[11pt]{article}

\usepackage{acl}

\usepackage{times}
\usepackage{latexsym}

\usepackage[T1]{fontenc}

\usepackage[utf8]{inputenc}

\usepackage{microtype}

\usepackage{inconsolata}

\usepackage{graphicx}

\usepackage{amsmath} 
\usepackage{amssymb}
\usepackage{algorithm}
\usepackage{algpseudocode}
\usepackage{todonotes}
\usepackage{lipsum}
\usepackage{booktabs}
\usepackage{paralist}
\usepackage{hyperref}
\usepackage{multirow}
\usepackage{makecell}
\usepackage{bm}
\usepackage{tabularx}
\newcolumntype{C}{>{\centering\arraybackslash}X}

\usepackage{pifont} %
\usepackage{xspace}
\usepackage[most]{tcolorbox}
\tcbuselibrary{breakable}
\DeclareTCBListing{promptbox}{ m }{
  colback=gray!5!white,
  colframe=gray!50!black,
  fonttitle=\bfseries\small,
  title={#1},
  breakable,
  left=4pt, right=4pt, top=4pt, bottom=4pt,
  listing only, %
  listing options={
    basicstyle=\ttfamily\footnotesize,
    breaklines=true, %
    columns=fullflexible, %
    showstringspaces=false
  }
}

\newcommand{\name}{\ensuremath{\mathsf{ABSOL}}\xspace}

\newcommand{\namenollm}{\ensuremath{\mathsf{ABSOL_{no LLM}}}\xspace}

\title{ABSOL: Aggregated Bayesian Subsampling Orchestrated with LLMs}

\author{Jackson Hassell \\
  Megagon Labs \\
  \texttt{jackson@megagon.ai} \\\And
  Chen Shen \\
  Megagon Labs \\
  \texttt{chen\_s@megagon.ai} \\ \And
  Estevam Hruschka \\
  Megagon Labs \\
  \texttt{estevam@megagon.ai} \\}

\begin{document}
\maketitle
\begin{abstract}
Large language models are increasingly used as natural-language interfaces to structured data, yet they remain unreliable when answers require consistent evidence conditioning, dependency-aware reasoning, and uncertainty estimation.
Bayesian networks provide an explicit probabilistic reasoning layer, but learning
useful structures from data remains costly and fragile at scale. We introduce \name{}, a hybrid LLM-guided Bayesian network structure-learning framework that
uses LLMs as bounded semantic guides.
Across five discrete BN benchmarks spanning 27 to 1041 nodes, \name{} is the only
evaluated method to produce a viable graph on every benchmark, and achieves the
highest Edge \(F_1\) on every benchmark larger than 27 nodes with GPT-5.4. 
The four LLM augmentations, which contribute complementary semantic evidence to the statistical backbone, improve Edge \(F_1\) over the non-LLM aggregation backbone by +0.23 on average.
Complementary post-hoc refinement experiments suggest that these gains depend in part on limiting the LLM's authority over the final structure.
Together, these results show that language-derived semantic knowledge can substantially improve scalable probabilistic structure learning when used as bounded guidance within a statistically grounded reasoning pipeline.
The code for \name is available at \url{github.com/megagonlabs/absol-bn}.
\end{abstract}

\section{Introduction}\label{sec-intro}

Bayesian networks (BNs) represent dependencies among variables as a
directed acyclic graph and support well-defined probabilistic inference
over evidence, targets, and latent relationships
\citep{pearl1988probabilistic,koller2009probabilistic}. Constructing useful BNs at realistic scale remains difficult, however: manual
specification is infeasible for domains with hundreds of variables, and purely data-driven structure learning is computationally expensive,
statistically fragile, and sensitive to finite-sample effects
\citep{scutari2019bigdata,kitson2023survey}.

At the same time, LLMs encode substantial semantic knowledge about how
domain variables tend to relate: which are plausibly associated, which
parent-child directions are semantically natural under a BN
interpretation, and which relationships are unlikely or only indirectly
mediated \citep{long2023can,darvariu2024llmpriors,ban2025harmonized}.
Recent NLP work on LLM-guided causal discovery and Bayesian Network structure elicitation either delegate graph construction directly to the
LLM or interleave LLM guidance with a full-data statistical learner
\citep{zhang2026bayesian,jiralerspong2024efficient,babakov2025scalability,long2023can,feng2025reliability,zhou2024causalbench,babakov2025causalgraphbench}, but
neither path has been shown to scale to probabilistic models with
hundreds of variables.

\begin{figure*}[h]
    \centering
    \includegraphics[width=0.99\textwidth, page=1]{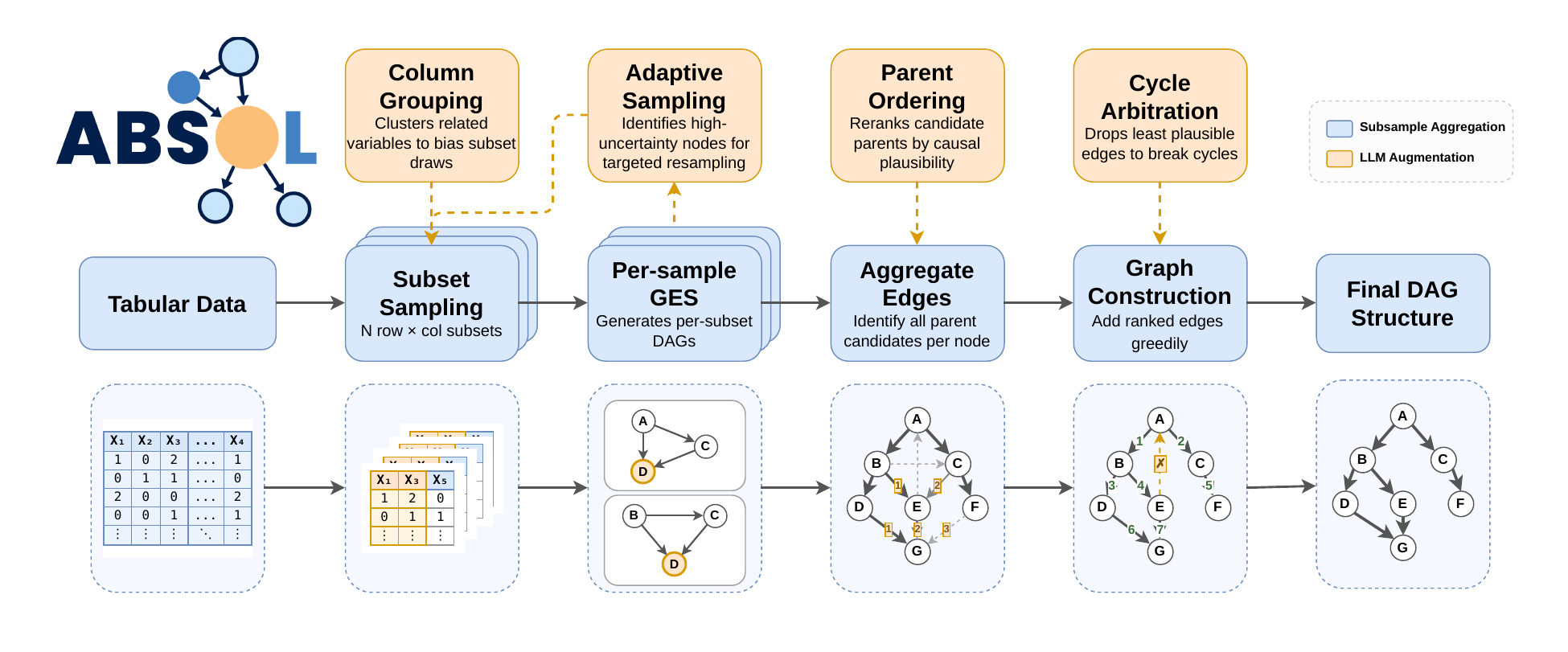}
    \caption{The \name{} algorithm. Blue components form the subsample-aggregated \textsc{GES+BDeu} backbone, and orange components are the four bounded LLM augmentations, each acting at a fixed decision point. Graph construction is specified fully in Algorithm \ref{alg:riolu}.}
    \label{fig:main-diagram}
\end{figure*}

We introduce \name{} (Figure~\ref{fig:main-diagram}), a hybrid LLM-guided BN structure-learning pipeline that scales to networks with hundreds of variables. Rather than asking an LLM to generate an entire graph, \name{} uses it only at bounded decision points, where semantic knowledge guides search while a statistical learner retains authority over structural decisions. 
This places \name{} within a broader NLP-systems pattern of using LLMs not as unconstrained predictors, but as semantic controllers inside verifiable, statistically grounded reasoning pipelines.

To improve scalability, \name{} learns BN subgraphs on many
row-and-column subsampled tables and aggregates their edge-level
decisions into a single global structure
(Section~\ref{sec-method-aggregation}), drawing on the idea that stable
structural signals can be recovered through repeated resampling and
aggregation
\citep{breiman1996bagging,friedman1999bootstrap,meinshausen2010stability,scutari2013identifying}.

To incorporate semantic knowledge and compensate for the reduced sample
size in each subsampled round, \name{} consults the LLM at four
bounded augmentations (Section~\ref{subsec-llmbag}): identifying variables
likely to interact (Column Grouping), focusing additional sampling on
ambiguous decisions (Adaptive Sampling), ranking candidate parents
(Parent Ordering), and resolving local acyclicity conflicts (Cycle
Arbitration). This design follows the emerging view that LLMs are more
reliable as bounded priors than as autonomous generators of complete
dependency graphs
\citep{darvariu2024llmpriors,ban2025harmonized,du2025llmcd}.
We evaluate \name{} on five discrete BN benchmarks ranging from 27 to
1041 nodes to test whether bounded LLM
guidance remains useful at scales where direct LLM graph generation and
full-data search become impractical.

Our results show that bounded LLM guidance substantially improves BN structure learning. Across benchmarks, \name{} achieves
the highest Edge \(F_1\) among the compared methods under GPT-5.4 at
every scale except \textsc{Insurance}. Under our evaluated implementations and computational budgets, no baseline produces a usable graph beyond \textsc{Neuropathic} (222 nodes), whereas \name{} runs successfully on \textsc{Diabetes} (413 nodes) and \textsc{Munin} (1041 nodes).
Ablations show that the four LLM-guided
components improve Edge \(F_1\) on every benchmark over \namenollm{}, with an average absolute lift of \(+0.23\) and the
largest relative gains on the largest graphs.
Conversely, the post-hoc refinement variants in Appendix~\ref{app:graph-refinement}, which give the LLM unilateral editing authority, reduce Edge \(F_1\) in eight of nine tested settings.

In summary, our contributions are threefold.
First, we provide an empirical account of how language-derived semantic knowledge can be integrated with statistical evidence, contrasting bounded LLM guidance with direct graph generation and post-hoc editing.
Second, we introduce a scalable subsample-aggregation backbone that decomposes
large BN structure learning into tractable local searches.
Third, we provide an empirical study of LLM semantic guidance at scale, showing
improvements across BN benchmarks up to 1041 nodes and isolating the contribution
of each bounded LLM augmentation.

\section{Related Work}\label{sec-relwork}

\subsection{Bayesian Network Structure Learning}

Bayesian network structure learning recovers a graph over variables from data \citep{pearl1988probabilistic,koller2009probabilistic}.
Existing methods fall into three families: constraint-based (e.g., PC), score-based (e.g., GES with BDeu scoring), and hybrid approaches \citep{kitson2023survey}.
For discrete data, \textsc{GES+BDeu} is the standard score-based baseline \citep{heckerman1995learning,chickering2002optimal}.
\name{} keeps \textsc{GES+BDeu} as the statistical core, but uses it through subsampled aggregation rather than a single full-data search.

\subsection{Scaling Structure Learning Through Aggregation}

Scaling structure learning to high-dimensional settings is challenging
both algorithmically and statistically. Prior work reduces per-instance
complexity by restricting the search space before learning -- via
candidate-parent restriction \citep{friedman1999sparsecandidate}, local
skeleton discovery combined with score search
\citep{tsamardinos2006maxmin}, greedy-search optimization for large
data \citep{scutari2019bigdata}, or learning and merging subgraphs from
separate variable-space partitions \citep{bernaola2023learning} -- at
the cost of potentially missing long-range dependencies.

A complementary framing asks which structural features persist under
data perturbations. Bootstrap aggregation \citep{breiman1996bagging},
applied to BN analysis, scores edge and Markov-blanket confidence by
learning networks across resampled tables, with follow-up work refining
model-averaging choices, replicate counts, and edge-confidence
thresholds
\citep{friedman1999bootstrap,broom2012model,scutari2013identifying}.
Stability selection generalizes this intuition through subsampling
followed by selection-frequency thresholds
\citep{meinshausen2010stability}.
\name{} extends this line by learning overlapping variable subproblems with \textsc{GES+BDeu} and aggregating edge decisions under an explicit finite-sample bound.

\subsection{LLM-Augmented Structure Learning for Bayesian Networks}

LLM-augmented causal discovery operates at multiple stages, with varying LLM roles in the final structure decision.
LLM-primary graph construction methods such as bfsBN \citep{jiralerspong2024efficient} use the LLM to directly construct candidate graph structure, reaching 222 nodes on Neuropathic. Hybrid methods instead constrain the LLM to guide rather than decide.
In-process guidance during algorithm execution appears most effective: score-based methods like Harmonized Prior \citep{ban2025harmonized} decompose LLM judgments into soft constraints for \textsc{GES+BDeu} scoring (reaching 109 nodes on Pathfinder), while constraint-based approaches like LLM-CD \citep{du2025llmcd} insert LLM guidance at skeleton construction and edge-orientation phases.
Post-learning verification applies LLM feedback after structure recovery: CauScientist \citep{peng2026causcientist} validates LLM proposals against data, and iterative refinement \citep{ban2023iterative} cycles between learning and LLM-guided improvement suggestions.
In the hybrid and in-process settings, a consistent principle holds: the data-driven learner (\textsc{GES+BDeu} or PC) retains final authority over structural decisions, with the LLM as a biasing signal rather than the decision mechanism. 
\name{} follows this guidance-based pattern, but applies LLM augmentation across a subsample-aggregation pipeline rather than a single full-data learner.

\section{ABSOL Method}\label{sec-method}

\subsection{Problem Definition and Standard BN Learning}
\label{sec-method-standard}

Let \(D=\{x^{(r)}\}_{r=1}^n\) be a discrete data table with \(n\) rows and variables \(X_1,\ldots,X_p\).
A Bayesian network (BN) consists of a directed acyclic graph \(G\) and local conditional distributions over these variables \citep{pearl1988probabilistic,koller2009probabilistic}.
Writing \(\mathrm{Pa}_i(G)\) for the parents of \(X_i\), the graph induces the factorization
\begin{equation}
\label{eq:bn-factorization}
P(X_1,\ldots,X_p)=\prod_{i=1}^p P(X_i\mid \mathrm{Pa}_i(G)).
\end{equation}
The structure-learning task is to estimate \(G\) from \(D\).

We use score-based BN structure learning as the statistical reference.
A score-based learner assigns a data-dependent score to each candidate graph and searches for a high-scoring structure.
For discrete variables, we use the Bayesian Dirichlet equivalent uniform (BDeu) score \citep{heckerman1995learning}.
Because BDeu decomposes across node families, local graph operations can be evaluated by rescoring only the affected child families.
We pair this score with Greedy Equivalence Search (GES), which greedily searches over Markov-equivalence classes of DAGs using local insert/delete operators \citep{chickering2002optimal}.

For any table \(D'\), whether the full data \(D\) or a row/variable subtable, let \(S_{\mathrm{BDeu}}(G;D')\) denote the BDeu score of graph \(G\) on \(D'\). We define the \textsc{GES+BDeu} base learner \(\mathcal A\) and its full-data reference output as
\begin{equation}
\label{eq:base-learner}
\begin{aligned}
\mathcal A(D')
&:=\textsc{GES}\!\big(D';S_{\mathrm{BDeu}}(\cdot;D')\big),\\
G_{\mathrm{std}}
&:=\mathcal A(D).
\end{aligned}
\end{equation}

Here \(G_{\mathrm{std}}\) denotes the graph returned by full-data \textsc{GES+BDeu}, which need not be the global maximizer of the BDeu score. The aggregation analysis below is therefore learner-relative: it asks when many smaller \(\mathcal A(D')\) runs reproduce the decisions of the full-data reference learner, rather than claiming recovery of the unknown causal graph.

\subsection{Subsample-Aggregated BN Learning}
\label{sec-method-aggregation}
Full-data \(\mathcal A\) is expensive because score-based BN search may need to enumerate parent/operator subsets, which is exponential in \(p\) in the worst case.
\name{} instead runs \(B\) independent searches on \(k\)-variable subtables, with total search cost \(O\!\left(B\,T_{\mathrm{GES}}(m,k)\right)\)
(Table~\ref{tab:baseline-comparison}).
The costly search therefore happens inside \(k\)-variable subproblems; \(p\) enters the aggregation budget through the endpoint co-sampling probability \(\pi_{\mathrm{elig}}=k(k{-}1)/(p(p{-}1))\).

In round \(b\), \name{} samples \(m\) rows and \(k\) variables uniformly without replacement, runs \(\mathcal A\) on the induced subtable, and embeds the learned graph \(\widehat G^{(b)}\) back into the full variable set.\footnote{Experiments include both this backbone and the LLM-augmented variants in Section~\ref{subsec-llmbag}; the bound here applies to the backbone.}

We aggregate the learned structures by \emph{co-occurrence voting}.
For an unordered edge \(e=\{u,v\}\) over the original \(p\) variables, let \(c_e\) denote the number of rounds in which both endpoints were co-sampled (\(\{u,v\}\subseteq C_b\)).
For any graph \(G\), let \(E_{\mathrm{skel}}(G)\) denote its unordered adjacency set.
Throughout this subsection, \(e\in G\) means \(e\in E_{\mathrm{skel}}(G)\), i.e., the two endpoints are adjacent regardless of direction.
Thus, the aggregation guarantee below concerns adjacency inclusion; directed assembly is handled downstream, with parent ordering and cycle arbitration providing the LLM-augmented cases in Section~\ref{subsec-llmbag}.
Define
\begin{equation}
\label{eq:edge-vote-frequency}
\widehat f_e
=
\begin{cases}
\frac{1}{c_e}
\sum_{b:\{u,v\}\subseteq C_b}
\mathbf{1}\{e\in \widehat G^{(b)}\},
& \text{if } c_e>0,\\
0,
& \text{if } c_e=0.
\end{cases}
\end{equation}
Rounds in which either endpoint is absent from \(C_b\) are excluded from both numerator and denominator; only rounds with genuine co-occurrence contribute.
On \(c_e>0\), \(\widehat f_e\) is an unbiased estimator of \(p_e^{\mathrm{co}}=\Pr[e\in\widehat G^{(b)}\mid\{u,v\}\subseteq C_b]\), the edge probability conditional on co-sampling.
Co-occurrence voting removes the mechanical shrinkage by the endpoint-eligibility rate \(\pi_{\mathrm{elig}}\); however, \(p_e^{\mathrm{co}}\) can still depend on the variable budget \(k\) through the other variables co-sampled in the subproblem.
The appendix derives the resulting concentration bound, including the case \(c_e=0\).
Given a threshold \(\tau\), the aggregated graph is
\begin{equation}
\label{eq:aggregated-graph}
\widehat G_B(\tau)=\{e:\widehat f_e\ge \tau\}.
\end{equation}

We next formalize two properties of the aggregated graph: it approximates the full-data reference learner on certifiable edge decisions, and its failure probability can be controlled by the number of subsampling rounds.

\noindent\textbf{Theorem (Aggregation approximation).}
Let \(G_{\mathrm{std}}\) be the full-data output of \(\mathcal A\) from Eq.~\ref{eq:base-learner}.
Let \(\mathcal M_{\mathrm{cert}}\) be the set of edges whose include/exclude verdict can be certified via the full-data GES operator trace; let \(M_{\mathrm{cert}}=|\mathcal M_{\mathrm{cert}}|\).
For each certified edge \(e\in\mathcal M_{\mathrm{cert}}^+\) (included in \(G_{\mathrm{std}}\)), define the positive margin \(a_e^+ = p_e^{\mathrm{co}}-\tau\); for \(e\in\mathcal M_{\mathrm{cert}}^-\) (excluded from \(G_{\mathrm{std}}\)), define the negative margin \(a_e^- = \tau-p_e^{\mathrm{co}}\).
Let \(a_{\min}=\min\left\{\min_{e\in\mathcal M_{\mathrm{cert}}^+}a_e^+, \min_{e\in\mathcal M_{\mathrm{cert}}^-}a_e^-\right\}\) be the weakest signed margin, with empty positive or negative classes omitted.
Under regularity, path-stability, and independent-round assumptions, if \(a_{\min}>0\), then
\begin{equation}
\label{eq:aggregation-approx-bound}
\begin{aligned}
&\Pr\!\left[
  \widehat{G}_B(\tau)\cap\mathcal M_{\mathrm{cert}}
  \ne
  G_{\mathrm{std}}\cap\mathcal M_{\mathrm{cert}}
\right] \\
&\qquad\le
M_{\mathrm{cert}}
\left(
  1-\pi_{\mathrm{elig}}\left(1-\exp(-2a_{\min}^2)\right)
\right)^B .
\end{aligned}
\end{equation}
The failure probability decays exponentially in \(B\); the certified-universe construction, path-stability conditions, and detailed proofs are deferred to Appendix~\ref{app:aggregation-proof}. The theorem characterizes the subsample-aggregation backbone, and it does not provide a practical certificate for a full \name{} run. Its practical role is to separate finite-vote disagreement, which additional rounds reduce, from limitations in each subsample's information that additional rounds cannot remove.

\noindent\textbf{Corollary (Aggregation budget).}
Since the realized margin \(a_{\min}\) is not known a priori, we fix a target resolvable margin \(a_{\mathrm{target}}\in(0,1/2]\) and a pre-specified upper bound \(M_0\le \binom{p}{2}\) on certified edges.
Applying the standard bound \((1-x)^B\le \exp(-Bx)\) to Eq.~\ref{eq:aggregation-approx-bound} at 5\% failure probability yields the sufficient design rule
\begin{equation}
\label{eq:b95-target-margin}
B_{95}(a_{\mathrm{target}}) = \left\lceil \frac{\log(20M_0)}{\pi_{\mathrm{elig}}(1-\exp(-2a_{\mathrm{target}}^2))} \right\rceil,
\end{equation}
where \(\pi_{\mathrm{elig}}=k(k{-}1)/(p(p{-}1))\) is the endpoint co-occurrence probability.
For small margins \(a_{\mathrm{target}} \ll 1\), the formula simplifies to the approximation
\[
B_{95}(a_{\mathrm{target}}) \approx
\left\lceil
\frac{\log(20M_0)}
{2\pi_{\mathrm{elig}}\,a_{\mathrm{target}}^2}
\right\rceil.
\]

We use \(a_{\mathrm{target}}=0.10\) as a reference margin for planning the subsample budget. Because \(a_{\min}\) is not known a priori, Eq.~\ref{eq:b95-target-margin} gives a conservative sufficient budget rather than a dataset-specific optimum. In the main experiments, we use a common experimental budget of \(B=1000\) (Table~\ref{tab:hyperparameters}) for all datasets without per-dataset tuning. Appendix~\ref{app:tau_ablation} reports how aggregation behaves as \(B\) varies, and Appendix~\ref{app:agg-diagnostics} lists the diagnostics for applying the bound.

\subsection{LLM Augmentation Strategies}
\label{subsec-llmbag}
The aggregated learner \(\widehat G_B(\tau)\) is grounded in data but blind to variable meaning.
\name{} consults an LLM through four bounded augmentations in the pipeline (Figure~\ref{fig:main-diagram}, Algorithm~\ref{alg:riolu}), using semantic knowledge while keeping \textsc{GES+BDeu} as the statistical backbone.
Each LLM response is parsed and validated before use, and if validation fails, the corresponding step falls back to its statistical baseline.
All LLM calls receive short natural-language descriptions of each variable and its discrete states (Appendix \ref{app:datasets-vardesc}).
Hyperparameters, prompt templates, and implementation details are deferred to Appendix~\ref{app:implementation}.

\noindent\textbf{Column Grouping}
Uniform variable subsampling can waste rounds on subsets whose variables are unlikely to interact.
For each \(X_i\), we ask the LLM which variables are plausible direct causes or effects, and symmetrize the responses into an adjacency graph \(\mathcal G_{\mathrm{LLM}}\).
During variable sampling, pairs adjacent in \(\mathcal G_{\mathrm{LLM}}\) are upweighted to co-occur more often, concentrating the fixed subsampling budget on semantically plausible dependencies.

\noindent\textbf{Adaptive Sampling}
After the initial \(B\) rounds, edges with vote frequencies near \(\tau\) are ambiguous; we score this ambiguity by the Bernoulli entropy of \(\widehat f_e\).
For nodes incident to high-entropy edges, the LLM is shown the uncertain candidates and their vote frequencies, then asked which additional variables should be co-sampled as possible confounders, mediators, or disambiguators.
Additional rounds force-include the suggested variables, and their edge votes and co-occurrence counts are merged with the initial pool (Algorithm~\ref{alg:adaptive}).

\noindent\textbf{Parent Ordering}
Raw vote frequency ranks association strength, not causal direction: direct causes can be mixed with confounded or transitively correlated variables.
For each node, the LLM sees every candidate parent with its subsampling support and returns a ranked list of plausible direct causes.
This resembles prior-evaluation approaches for LLM causal knowledge~\citep{darvariu2024llmpriors}, but the ranking is used inside the aggregation pipeline rather than supplied as external constraints to a separate discovery algorithm.

\noindent\textbf{Cycle Arbitration}
Unlike broader LLM-based graph cleanup or refinement \citep{du2025llmcd,ban2023iterative}, \name{} invokes the LLM only when greedy assembly would violate acyclicity, limiting semantic intervention to local cases where the statistical assembler would otherwise reject a candidate edge.
When a candidate edge \(P\to C\) would close one or more cycles, we enumerate the directed paths \(C\leadsto P\) in the graph and ask the LLM for a subset of path edges whose removal breaks every offending path.

We also evaluated a fifth augmentation, post-hoc graph refinement in the style of prior LLM-based refinement work~\citep{ban2023iterative}.
Because this variant consistently degraded performance, we exclude it from \name{} and report the structural failure mode in Appendix~\ref{app:graph-refinement}.

\subsection{Augmentation Assembly}
\label{subsec-assembly}

The four LLM augmentations intervene at distinct stages of the \name pipeline. Before structure learning begins, column grouping constructs a semantic adjacency graph that biases the variable-subset sampler toward placing plausibly related variables in the same subproblems. \name then performs the initial row-and-column subsampling rounds, runs \textsc{GES+BDeu} on each induced subtable, and aggregates the resulting directed-edge observations using co-occurrence-normalized support. Adaptive sampling is applied after this initial aggregation stage: directed edges with sufficiently high Bernoulli entropy are treated as uncertain, and the LLM proposes additional variables to co-sample with the affected nodes. The resulting targeted rounds are merged with the initial rounds by updating the same edge-vote and co-occurrence counts.

After all initial and adaptive rounds are complete, \name converts the final aggregated evidence into a DAG. For each child $X_i$, it forms a candidate-parent list from variables observed as parents of $X_i$ in the learned subgraphs, together with their co-occurrence-normalized support. Parent ordering reranks this list using both the subsampling evidence and the variables' semantic descriptions. LLM-endorsed candidates may remain eligible below the aggregation threshold $\tau$, whereas non-endorsed candidates must clear $\tau$. The assembler then processes candidate edges greedily in this order, adding $X_j \rightarrow X_i$ only while respecting the maximum in-degree and acyclicity constraints. If an otherwise eligible edge would create a cycle, cycle arbitration is invoked on the existing paths from $X_i$ to $X_j$: the LLM proposes a set of path edges whose removal would break every resulting cycle, and the proposed edit is applied only if it passes structural validation. The complete procedure is given in Algorithm~\ref{alg:riolu}.

\section{Experimental Evaluation}\label{sec-exp}

We organize the experiments around three questions: (i) how the
subsample-aggregated \textsc{GES+BDeu} backbone compares with the
full-data learner where feasible,
(ii) which LLM augmentations improve BN structure learning, and
what this can tell us about combining LLM and statistical evidence, 
and (iii) whether \name{} remains 
accurate and tractable as graph size increases.

\begin{table}[t]
\centering
\small
\begin{tabular}{@{}lccc@{}}
    \toprule
    \textbf{Name} & \makecell{\textbf{\# Nodes}} & \makecell{\textbf{\# Edges}} & \textbf{Domain} \\
    \midrule
    \textsc{Insurance}   & 27   & 52   & Finance \\
    \textsc{Hepar2}      & 70   & 123  & Medical \\
    \textsc{Neuropathic} & 222  & 770  & Medical \\
    \textsc{Diabetes}    & 413  & 602  & Medical \\
    \textsc{Munin}       & 1041 & 1397 & Medical \\
    \bottomrule
\end{tabular}
\caption{Chosen Bayesian network benchmark details.}
\label{tab:networks}
\end{table}

\begin{table}[t]
\centering
\tiny
\setlength{\tabcolsep}{3pt}
\renewcommand{\arraystretch}{1.1}
\begin{tabular}{l c l c}
\toprule
\textbf{Method} & \textbf{Type} & \textbf{Non-LLM Work} & \textbf{LLM Calls} \\
\midrule
\textsc{GES+BDeu} & Stat. & $T_{\mathrm{GES}}(n,p)$ & 0 \\
\textsc{PC}       & Stat. & $T_{\mathrm{PC}}(n,p,d)$ & 0 \\
\textsc{LLM-CD}   & Hyb.  & $O(I\,T_{\mathrm{PC}}(n,p,d))$ & $O(I p^2 2^d)$ \\
\textsc{PromptBN} & LLM   & --- & 1 \\
\textsc{bfsBN}    & LLM   & --- & $O(p)$ \\
\namenollm{}      & Hyb.  & $O(B\,T_{\mathrm{GES}}(m,k))$ & 0 \\
\name{}           & Hyb.  & $O((B+t|\mathcal{U}|)\,T_{\mathrm{GES}}(m,k))$ & $O(p+|\mathcal{U}|+C)$ \\
\bottomrule
\end{tabular}
\caption{Per-method comparison of \name{} and the baselines.
\emph{Non-LLM Work} is the sequential structure-search cost (excluding
CPT fitting), \emph{LLM Calls} counts API requests. Symbols,
parallelism notes, and derivations are given in
Appendix~\ref{app:cost-derivations}.}
\label{tab:baseline-comparison}
\end{table}

\subsection{Datasets}\label{subsec-datasets}

We evaluate on five discrete Bayesian network benchmarks spanning two orders of
magnitude in graph size: \textsc{Insurance}, \textsc{Hepar2},
\textsc{Neuropathic}, \textsc{Diabetes}, and \textsc{Munin}.
All but \textsc{Neuropathic} are drawn from the standard \texttt{bnlearn}
repository \citep{scutari2010learning}, which supplies the canonical benchmark
suite for prior LLM-based BN structure discovery
\citep{babakov2025scalability,jiralerspong2024efficient,zhang2026bayesian}.
\textsc{Neuropathic} \citep{tu2019neuropathic} is a synthetic
clinical network used in prior LLM-driven causal discovery work \citep{jiralerspong2024efficient}. For each
ground truth network, we generate a synthetic discrete training table (see Appendix~\ref{app:datasets} for details)
and evaluate the learned structure from the synthetic data against that same network.

\subsection{Baselines}\label{subsec-baselines}

\begin{table*}[t]
\centering
\small
\resizebox{\textwidth}{!}{%
\setlength{\tabcolsep}{3pt}%
\begin{tabular}{lccccccccccccccc}
\toprule
 & \multicolumn{3}{c}{\textbf{Insurance}} & \multicolumn{3}{c}{\textbf{Hepar2}} & \multicolumn{3}{c}{\textbf{Neuropathic}} & \multicolumn{3}{c}{\textbf{Diabetes}} & \multicolumn{3}{c}{\textbf{Munin}} \\
\cmidrule(lr){2-4}\cmidrule(lr){5-7}\cmidrule(lr){8-10}\cmidrule(lr){11-13}\cmidrule(lr){14-16}
Method & NHD $\downarrow$ & Edge \(F_1\) $\uparrow$ & BDeu $\uparrow$ & NHD $\downarrow$ & Edge \(F_1\) $\uparrow$ & BDeu $\uparrow$ & NHD $\downarrow$ & Edge \(F_1\) $\uparrow$ & BDeu $\uparrow$ & NHD $\downarrow$ & Edge \(F_1\) $\uparrow$ & BDeu $\uparrow$ & NHD $\downarrow$ & Edge \(F_1\) $\uparrow$ & BDeu $\uparrow$ \\
\midrule
GES & \textbf{0.017} & 0.920 & $\bm{-2.6\mathrm{e}{6}}$ & \multicolumn{3}{c}{OoM} & \multicolumn{3}{c}{OoM} & \multicolumn{3}{c}{OoM} & \multicolumn{3}{c}{OoM} \\
PC & 0.063 & 0.674 & $-2.8\mathrm{e}{6}$ & 0.037 & 0.391 & $-6.6\mathrm{e}{6}$ & \textbf{0.033} & 0.135 & $-6.7\mathrm{e}{6}$ & \multicolumn{3}{c}{ToF} &\multicolumn{3}{c}{ToF} \\
PromptBN & 0.074 & 0.764 & $-2.8\mathrm{e}{6}$ & \multicolumn{3}{c}{OoT} & \multicolumn{3}{c}{OoT} & \multicolumn{3}{c}{NVG} & \multicolumn{3}{c}{OoT} \\
bfsBN & 0.060 & 0.774 & $-2.9\mathrm{e}{6}$ & 0.058 & 0.368 & $-6.7\mathrm{e}{6}$ & \multicolumn{3}{c}{OoC} & \multicolumn{3}{c}{OoT} & \multicolumn{3}{c}{OoC} \\
LLM-CD & {0.017} & \textbf{0.939} & $-2.8\mathrm{e}{6}$ & 0.019 & 0.751 & \bm{$-6.5\mathrm{e}{6}$} & \multicolumn{3}{c}{ToF} & \multicolumn{3}{c}{ToF} & \multicolumn{3}{c}{ToF} \\
\namenollm{} & 0.048 & 0.796 & \bm{$-2.6\mathrm{e}{6}$} & 0.051 & 0.493 & \bm{$-6.5\mathrm{e}{6}$} & 0.039 & 0.256 & $-6.3\mathrm{e}{6}$ & 0.015 & 0.073 & $-1.2\mathrm{e}{8}$ & 0.006 & 0.025 & $-9.9\mathrm{e}{7}$ \\
\name & {0.023} & {0.902} & \bm{{$-2.6\mathrm{e}{6}$}} & \textbf{0.017} & \textbf{0.798} & \bm{$-6.5\mathrm{e}{6}$} & \textbf{0.033} & \textbf{0.423} & \bm{$-5.8\mathrm{e}{6}$} & \textbf{0.007} & \textbf{0.499} & \bm{$-7.0\mathrm{e}{7}$} & \textbf{0.005} & \textbf{0.187} & \bm{$-8.6\mathrm{e}{7}$} \\
\bottomrule
\end{tabular}
}
\caption{Each baseline and \name (with and without LLM augmentations) across five Bayesian network benchmarks. LLM-powered methods use GPT-5.4, and each cell is the result of a single run. See Table~\ref{tab:augmentation-ablation} for multiple trials and Table~\ref{tab:llm-ablation} for cross-LLM comparison. Best per column in bold. Methods that failed to finish within resource limits are annotated with their failure reason: OoM=out of memory failure, OoT=out of per-response tokens failure, OoC=out of context (context window filled) failure, ToF=timeout failure, NVG=no valid graph, such as not conforming to the output format expected or returning a template that was not evaluable as a DAG.}
\label{tab:main-results}
\end{table*}

We compare \name against two purely statistical
structure learners, three recent LLM-driven approaches, and a purely statistical variant of the \name pipeline. Each baseline is reimplemented to match the configuration reported in its source paper, including hyperparameters, depth caps, and scoring choices. Per-baseline implementation details are deferred to Appendix~\ref{app:implementation}.
\begin{itemize}
\setlength{\itemsep}{2pt}
\item \textsc{GES} \citep{chickering2002optimal} The score-based learner
that \name uses internally as its base learner. Running it directly on the full
table isolates the contribution of subsample aggregation.

\item \textsc{PC} \citep{spirtes2000causation} The canonical
constraint-based learner, providing a non-score-based statistical reference
point.

\item \textsc{PromptBN} \citep{zhang2026bayesian} Generates a full DAG in a
single LLM call from variable metadata alone, representing the data-free end of
the spectrum.

\item \textsc{bfsBN} \citep{jiralerspong2024efficient} Constructs a DAG via
a breadth-first multi-turn LLM conversation, representing
LLM-as-structure-learner approaches that use no tabular data.

\item \textsc{LLM-CD} \citep{du2025llmcd} Hybrid approach that injects LLM
guidance into PC's skeleton-discovery and edge-orientation phases,
representing in-process LLM-as-guide approaches that interleave LLM judgments
with a constraint-based learner on the full data table.

\item \namenollm{}: The subsample-aggregated \textsc{GES+BDeu} pipeline of Section~\ref{sec-method-aggregation}, with no LLM augmentations.
\end{itemize}

Table~\ref{tab:baseline-comparison} summarizes each method's structure-search cost and LLM call count.

\subsection{Evaluation Metrics}\label{subsec-metrics}

We report three structure-quality measures against the ground-truth
network:

\textbf{Normalized Hamming Distance (NHD)}: fraction of ordered variable pairs whose directed
adjacency differs from ground truth, normalized by \(p(p-1)\) (lower is
better). 

\textbf{Edge \(\mathbf{F_1}\)}: standard F-measure on directed edges, with
precision \(|E_{\text{learned}}\cap E_{\text{gt}}|/|E_{\text{learned}}|\)
and recall \(|E_{\text{learned}}\cap E_{\text{gt}}|/|E_{\text{gt}}|\)
(higher is better).

\textbf{BDeu} \citep{heckerman1995learning}:
Bayesian Dirichlet equivalent uniform score of the learned structure on
the training table (higher is better).

\subsection{Main Results}\label{subsec-main-results}
Table~\ref{tab:main-results} shows the main comparison across the five benchmarks. \name{} is the only evaluated method that produces a viable graph on every benchmark and obtains the highest Edge \(F_1\) on every benchmark larger than \textsc{Insurance}. On \textsc{Insurance}, full-data \textsc{GES} and \textsc{LLM-CD} outscore \name{} because the \(p{=}27\) graph is still small enough for monolithic search, giving each structural decision access to the full variable set. This monolithic strategy becomes infeasible as graphs grow: \textsc{GES} exhausts memory by \textsc{Hepar2}, and \textsc{LLM-CD} times out by \textsc{Neuropathic}. In contrast, \name{} keeps each structure-learning subproblem fixed-size and uses LLM calls that scale linearly in \(p\). Even on \textsc{Hepar2}, where \textsc{LLM-CD} still completes, \name{} reaches comparable Edge \(F_1\) in 20 minutes rather than 30 hours (Tables~\ref{tab:main-results},~\ref{tab:cost}).

The baseline failures are also informative. Full-data \textsc{GES} exhausts memory on \textsc{Hepar2}, while PC completes \textsc{Neuropathic} only after 35 hours and then times out on larger networks. \textsc{PromptBN} usually exceeds its 16K-token completion budget, but on \textsc{Diabetes} it instead returns an invalid dynamic-BN template rather than an evaluable DAG. \textsc{bfsBN} exhausts GPT-5.4's context window during the \textsc{Neuropathic} traversal as conversation history accumulates, and exceeds the token budget on the first call for \textsc{Diabetes}.

\textsc{LLM-CD} completes \textsc{Hepar2}, but does not finish on \textsc{Neuropathic} within 48 hours because its skeleton phase is neither depth-bounded nor parallelized. Appendix~\ref{app:implementation} gives the full mechanisms.

The comparison between \namenollm{} and \name{} isolates the contribution of the LLM-guided components. Without any LLM calls, \namenollm{} already beats \textsc{PC}, \textsc{PromptBN}, and \textsc{bfsBN} on Edge \(F_1\) wherever those baselines complete. Adding the four LLM augmentations from Section~\ref{subsec-llmbag} improves Edge \(F_1\) on every benchmark, with an average absolute lift of \(+0.23\) over \namenollm{}. The relative lift is largest on the largest graph, \textsc{Munin}, where Edge \(F_1\) rises more than seven-fold (\(0.025\!\to\!0.187\)). On the three benchmarks with five-trial estimates, \name{}'s 95\% Edge \(F_1\) confidence intervals are within \(0.020\), several times smaller than the lifts over \namenollm{} (Table~\ref{tab:augmentation-ablation}). BDeu and NHD follow the same overall pattern, and \name{} improves over \namenollm{} across the reported metrics and benchmarks. This pattern is consistent with the intended role of semantic guidance: as the candidate-edge pool grows, the subsampling signal becomes sparser and domain knowledge has greater marginal value.

\begin{table*}[t]
\centering
\small
\resizebox{\textwidth}{!}{%
\setlength{\tabcolsep}{3pt}%
\begin{tabular}{llccccccccccccccc}
\toprule
 &  & \multicolumn{3}{c}{\textbf{Insurance}} & \multicolumn{3}{c}{\textbf{Hepar2}} & \multicolumn{3}{c}{\textbf{Neuropathic}} & \multicolumn{3}{c}{\textbf{Diabetes}} & \multicolumn{3}{c}{\textbf{Munin}} \\
\cmidrule(lr){3-5}\cmidrule(lr){6-8}\cmidrule(lr){9-11}\cmidrule(lr){12-14}\cmidrule(lr){15-17}
Method & LLM & NHD $\downarrow$ & Edge \(F_1\) $\uparrow$ & BDeu $\uparrow$ & NHD $\downarrow$ & Edge \(F_1\) $\uparrow$ & BDeu $\uparrow$ & NHD $\downarrow$ & Edge \(F_1\) $\uparrow$ & BDeu $\uparrow$ & NHD $\downarrow$ & Edge \(F_1\) $\uparrow$ & BDeu $\uparrow$ & NHD $\downarrow$ & Edge \(F_1\) $\uparrow$ & BDeu $\uparrow$ \\
\midrule
PromptBN & GPT & 0.074 & 0.764 & $-2.8\mathrm{e}{6}$ & \multicolumn{3}{c}{OoT} & \multicolumn{3}{c}{OoT} & \multicolumn{3}{c}{NVG} & \multicolumn{3}{c}{OoT} \\
 & DeepSeek & \multicolumn{3}{c}{NVG} & \multicolumn{3}{c}{OoT} & \multicolumn{3}{c}{OoT} & \multicolumn{3}{c}{OoT} & \multicolumn{3}{c}{NVG} \\
 & Gemini & \textbf{0.043} & \textbf{0.845} & $\bm{-2.7\mathrm{e}{6}}$ & \textbf{0.053} & \textbf{0.429} & $\bm{-6.8\mathrm{e}{6}}$ & \multicolumn{3}{c}{OoT} & \multicolumn{3}{c}{OoT} &\multicolumn{3}{c}{NVG} \\
\midrule
bfsBN & GPT & 0.060 & 0.774 & $-2.9\mathrm{e}{6}$ & 0.058 & 0.368 & $\bm{-6.7\mathrm{e}{6}}$ &\multicolumn{3}{c}{OoC} &\multicolumn{3}{c}{OoT} & \multicolumn{3}{c}{OoC} \\
 & DeepSeek  & \multicolumn{3}{c}{OoT} & \multicolumn{3}{c}{OoT} & \textbf{0.031} & \textbf{0.171} & $\bm{-6.5\mathrm{e}{6}}$ & \multicolumn{3}{c}{OoC} & \multicolumn{3}{c}{OoT} \\
 & Gemini & \textbf{0.028} & \textbf{0.900} & $\bm{-2.7\mathrm{e}{6}}$ & \textbf{0.039} & \textbf{0.427} & $-6.8\mathrm{e}{6}$ & \multicolumn{3}{c}{OoC} & \multicolumn{3}{c}{OoC} & \multicolumn{3}{c}{OoT} \\
\midrule
LLM-CD & GPT & {0.017} & {0.939} & ${-2.8\mathrm{e}{6}}$ & 0.019 & 0.751 & \bm{$-6.5\mathrm{e}{6}$} & \multicolumn{3}{c}{ToF} & \multicolumn{3}{c}{ToF} & \multicolumn{3}{c}{ToF} \\
 & DeepSeek & 0.017 & 0.931 & $\bm{-2.6\mathrm{e}{6}}$ & 0.019 & 0.751 & ${\bm{-6.5\mathrm{e}{6}}}$ & \multicolumn{3}{c}{ToF} & \multicolumn{3}{c}{ToF} & \multicolumn{3}{c}{ToF} \\
 & Gemini & \textbf{0.014} & \textbf{0.949} & ${-2.6\mathrm{e}{6}}$ & 0.014 & \textbf{0.817} & ${\bm{-6.5\mathrm{e}{6}}}$ & \multicolumn{3}{c}{ToF} & \multicolumn{3}{c}{ToF} & \multicolumn{3}{c}{ToF} \\
\midrule
\name{} & GPT & 0.023 & 0.902 & $-2.6\mathrm{e}{6}$ & \textbf{0.017} & \textbf{0.798} & $\bm{-6.5\mathrm{e}{6}}$ & \textbf{0.033} & 0.423 & $\bm{-5.8\mathrm{e}{6}}$ & 0.007 & 0.499 & $-7.0\mathrm{e}{7}$ & 0.005 & 0.187 & $-8.6\mathrm{e}{7}$ \\
 & DeepSeek & 0.051 & 0.804 & $-2.6\mathrm{e}{6}$ & 0.018 & 0.761 & $\bm{-6.5\mathrm{e}{6}}$ & 0.035 & 0.380 & $-5.9\mathrm{e}{6}$ & 0.008 & 0.446 & $-7.4\mathrm{e}{7}$ & 0.005 & 0.184 & $-8.5\mathrm{e}{7}$ \\
 & Gemini & \textbf{0.017} & \textbf{0.922} & $\bm{-2.6\mathrm{e}{6}}$ & \textbf{0.017} & 0.789 & $\bm{-6.5\mathrm{e}{6}}$ & 0.034 & \textbf{0.460} & $-5.9\mathrm{e}{6}$ & \textbf{0.006} & \textbf{0.562} & $\bm{-6.4\mathrm{e}{7}}$ & \textbf{0.005} & \textbf{0.221} & $\bm{-8.4\mathrm{e}{7}}$ \\
\bottomrule
\end{tabular}
}
\caption{LLM backend comparison on 5 datasets. Models used: \texttt{gpt-5.4-2026-03-05, deepseek-v4-pro, gemini-3.1-pro-preview}. Bold marks the best LLM for each method on each dataset/metric. Failure codes as in Table~\ref{tab:main-results}.}
\label{tab:llm-ablation}
\end{table*}

We next consider whether the gains could be explained by benchmark memorization rather than useful semantic guidance.
Public Bayesian network benchmarks may be partially recoverable from variable names alone, and recent causal-discovery prompting studies document such memorization effects~\citep{long2023can,feng2025reliability,babakov2025causalgraphbench}.
Two comparisons argue against memorization as the main explanation.
First, the lift from \namenollm{} to \name{} remains large on the two largest benchmarks, where recalling an entire graph from pretraining is least plausible: \(+0.43\) Edge \(F_1\) on \textsc{Diabetes} (\(p=413\)) and \(+0.16\) on \textsc{Munin} (\(p=1041\)).
Second, on \textsc{Insurance}, where memorization is most plausible, the metadata-only \textsc{PromptBN}+Gemini baseline reaches Edge \(F_1=0.845\), while \name{}+Gemini reaches \(0.922\) on the same benchmark (Table~\ref{tab:llm-ablation}).
This gap suggests that the data-driven learner adds signal beyond name-based graph recall.
Moreover, \textsc{PromptBN}'s recall drops sharply after \textsc{Insurance} and does not scale beyond \textsc{Hepar2}, while \name{} continues to improve over its no-LLM backbone on the largest graphs.

Finally, the augmentation effect is not tied to a single LLM backend. All 15 (LLM, dataset) combinations in Table~\ref{tab:llm-ablation} improve over \namenollm{} in Edge \(F_1\), with mean lifts of \(+0.23\) for GPT-5.4, \(+0.19\) for DeepSeek, and \(+0.26\) for Gemini.
The same baseline comparison also holds across backends: with the same LLM, \name{} outperforms the LLM-powered baselines except for \textsc{LLM-CD} on \textsc{Insurance}, and for \textsc{LLM-CD}+Gemini on \textsc{Hepar2}.
Together, these results suggest that the gains come from the bounded integration of semantic augmentation into the subsample-aggregated learner, rather than from one particular model.

\subsection{LLM Augmentation Ablation Results}\label{subsec-results-augmentations}

\begin{table}[t]
\centering
\tiny
\setlength{\tabcolsep}{1pt}
\begin{tabular}{lcccc}
\toprule
Method & \textbf{Insurance} & \textbf{Hepar2} & \textbf{Neuropathic} & \textbf{Rank} \\
\midrule
\multicolumn{5}{l}{\textit{Edge \(F_1\)}} \\
\midrule
\namenollm{} & $0.789 \pm 0.015$ & $0.521 \pm 0.016$ & $0.222 \pm 0.020$ & $6.00$ \\
\name{} & $0.896 \pm 0.008$ & $0.772 \pm 0.019$ & $0.400 \pm 0.015$ & $\mathbf{2.00}$ \\
-- ParOrd & ${0.843 \pm 0.007}$ & ${0.732 \pm 0.010}$ & ${0.341 \pm 0.012}$ & ${4.67}$ \\
-- ColGroup & $0.874 \pm 0.006$ & $0.752 \pm 0.010$ & $0.395 \pm 0.020$ & $3.33$ \\
-- CycleArb & $\mathbf{0.898 \pm 0.013}$ & $\mathbf{0.781 \pm 0.015}$ & $0.381 \pm 0.006$ & $\mathbf{2.00}$ \\
-- AdaSam & $0.886 \pm 0.006$ & $0.615 \pm 0.005$ & $\mathbf{0.427 \pm 0.003}$ & $3.00$ \\
\midrule
\multicolumn{5}{l}{\textit{NHD}} \\
\midrule
\namenollm{} & $0.051 \pm 0.003$ & $0.048 \pm 0.002$ & $0.047 \pm 0.001$ & $6.00$ \\
\name{} & $\mathbf{0.025 \pm 0.002}$ & $0.019 \pm 0.002$ & $\mathbf{0.034 \pm 0.001}$ & $\mathbf{1.33}$ \\
-- ParOrd & ${0.038 \pm 0.001}$ & ${0.020 \pm 0.001}$ & ${0.035 \pm 0.001}$ & ${3.33}$ \\
-- ColGroup & $0.031 \pm 0.002$ & $0.020 \pm 0.001$ & $0.035 \pm 0.001$ & $3.00$ \\
-- CycleArb & $\mathbf{0.025 \pm 0.003}$ & $\mathbf{0.017 \pm 0.001}$ & $0.035 \pm 0.000$ & $\mathbf{1.33}$ \\
-- AdaSam & $0.029 \pm 0.002$ & $0.039 \pm 0.001$ & $0.044 \pm 0.001$ & $4.33$ \\
\midrule
\multicolumn{5}{l}{\textit{BDeu ($\times 10^{6}$)}} \\
\midrule
\namenollm{} & $-2.644 \pm 0.009$ & $-6.520 \pm 0.004$ & $-6.608 \pm 0.183$ & $6.00$ \\
\name{} & $\mathbf{-2.619 \pm 0.001}$ & $\mathbf{-6.507 \pm 0.002}$ & $-5.859 \pm 0.082$ & $\mathbf{1.67}$ \\
-- ParOrd & ${-2.624 \pm 0.003}$ & ${-6.509 \pm 0.002}$ & $\mathbf{-5.791 \pm 0.022}$ & ${2.67}$ \\
-- ColGroup & $-2.620 \pm 0.001$ & $-6.513 \pm 0.004$ & ${-5.852 \pm 0.059}$ & $3.33$ \\
-- CycleArb & $-2.620 \pm 0.001$ & $-6.511 \pm 0.006$ & $-5.863 \pm 0.037$ & $3.33$ \\
-- AdaSam & $\mathbf{-2.619 \pm 0.001}$ & $-6.512 \pm 0.003$ & $-6.372 \pm 0.025$ & $3.33$ \\
\end{tabular}
\caption{\name{} leave-one-out ablation to isolate the effects of each augmentation. All metrics are reported averaged across five trials (mean $\pm$ 95\% CI). Best mean per column in bold. All LLM-based methods use GPT-5.4. Rank is averaged for each method across all datasets, per metric.}
\label{tab:augmentation-ablation}
\end{table}

Table~\ref{tab:augmentation-ablation} reports leave-one-out (LOO) ablations for the four LLM augmentations on the three datasets where repeated runs are computationally feasible.
The full \name{} pipeline improves Edge \(F_1\) over \namenollm{} by \(+0.11\) on \textsc{Insurance}, \(+0.25\) on \textsc{Hepar2}, and \(+0.18\) on \textsc{Neuropathic}. These gaps are several times larger than the corresponding 95\% confidence intervals.
The ablation therefore studies which augmentations account for the gain.

\emph{Parent ordering} provides the most consistent Edge \(F_1\) benefit.
Removing it lowers Edge \(F_1\) on all three datasets, producing the largest drop on \textsc{Insurance} and \textsc{Neuropathic}, and the second-largest drop on \textsc{Hepar2}.
This suggests that statistical aggregation recovers useful adjacencies, while semantic directionality is important for converting them into parent sets.

\emph{Adaptive sampling} helps when uncertain edges can be resolved by targeted co-sampling, but its effect is dataset-dependent.
Removing it gives the largest Edge \(F_1\) drop on \textsc{Hepar2} and a smaller drop on \textsc{Insurance}, showing that extra rounds around high-entropy edges can improve the aggregated graph.
On \textsc{Neuropathic}, however, removing adaptive sampling slightly improves Edge \(F_1\) (\(0.400\to0.427\)).
This exception is consistent with the dataset's many left/right variable pairs: targeted resampling can concentrate on ambiguous near-duplicates, whereas uniform sampling may provide stronger smoothing.
The same variant worsens NHD and BDeu on \textsc{Neuropathic}, so the Edge \(F_1\) gain is not a uniform improvement.

\emph{Cycle arbitration} is a scale-dependent safeguard rather than a frequent intervention on the repeated-run datasets.
It is rarely invoked on smaller graphs: during \textsc{Hepar2} training it fires only once, and its leave-one-out effect is within the confidence interval.
On larger assembled graphs, cycle-closing conflicts are more common: on \textsc{Munin}, the same mechanism fires 62 times.
The measurable \textsc{Neuropathic} drop when cycle arbitration is removed (\(0.400\to0.381\)) is consistent with this pattern.
We therefore retain it because it is nearly inactive when unnecessary but provides a bounded way to resolve acyclicity conflicts when graph assembly becomes harder.

\emph{Column grouping} has the smallest standalone leave-one-out effect, but the LOO mean remains below \name{} on all three datasets. Its contribution appears partly entangled with later stages: among edges present with all augmentations but absent without column grouping, \(85\%\) on \textsc{Hepar2} and \(82\%\) on \textsc{Neuropathic} are also lost when parent ordering or cycle arbitration is removed. This overlap suggests that column grouping is associated with edges that are jointly supported across multiple augmentation stages.

Overall, the ablation supports keeping the full augmentation set.
Parent ordering provides the most consistent Edge \(F_1\) benefit. Removing it lowers Edge \(F_1\) on every dataset, with non-overlapping reported confidence intervals, and gives the worst average Edge-\(F_1\) rank among the leave-one-out variants.
Adaptive sampling produces the largest single-dataset effect on \textsc{Hepar2}, but is more dataset-dependent.
The remaining augmentations have smaller or more regime-specific effects, with each contributing evidence somewhere in the table or in the call-frequency analysis above.
The rank column gives the same summary: the full pipeline has the best or tied-best average rank for every metric, and the best overall average rank across metrics.
We therefore use all four augmentations as the default \name{} configuration.
\section{Conclusion}\label{sec-conclusion}

We introduced \name{}, a hybrid Bayesian network structure-learning pipeline that pairs subsample-aggregated \textsc{GES+BDeu} with four bounded LLM augmentations: column grouping, adaptive sampling, parent ordering, and cycle arbitration. The aggregation step is supported by a finite-sample bound against a full-data reference learner, while the LLM augmentations inject semantic guidance through constrained, graph-validating operations.

Our experiments show that this design changes the scalability frontier for LLM-assisted BN learning.
Every evaluated baseline fails by \textsc{Diabetes} (413 nodes), whereas \name{} produces viable graphs on both \textsc{Diabetes} and \textsc{Munin} (1041 nodes).
Bounded LLM guidance improves Edge \(F_1\) on every benchmark, with an average absolute lift of \(+0.23\) over \namenollm{} and the largest relative gains on the largest graphs; the pattern is consistent across three LLMs and five datasets.
The ablations further show that the most reliable LLM contributions come from targeted semantic interventions, especially parent ordering, rather than direct graph generation or unconstrained graph editing.

\section*{Limitations}\label{sec:limitations}

\noindent\textbf{Scope of the aggregation guarantee.}
The finite-sample guarantee in Section~\ref{sec-method-aggregation} isolates the error introduced by subsampling: it bounds disagreement between the aggregated graph and the full-data \textsc{GES+BDeu} reference learner on certifiable adjacency decisions.
It does not separately establish causal consistency of the full-data learner.
As with standard BN structure learning, performance can be affected by Markov-equivalence ambiguity, BDeu prior choices, finite-sample effects, and unobserved confounding.

\noindent\textbf{Benchmark setting.}
Following standard BN evaluation practice, the training tables are forward-sampled from known reference graphs.
This enables controlled comparison across methods, but does not test measurement noise, distribution shift, selection bias, or latent confounding in real observational data.
The benchmark graphs are also public, so LLMs may have seen variable names or graph fragments during pretraining.
Because benchmark leakage cannot be ruled out completely, Section~\ref{subsec-main-results} checks whether metadata-only baselines and scaling behavior can explain the observed gains.

\noindent\textbf{Contamination risks.}
Contamination-robust evaluation is an open question for LLM-guided causal structure learning. The Bayesian networks in the bnlearn repository \cite{scutari2010learning} are widely used because they have been curated and verified by human experts, but that maturity makes pretraining exposure likely. The cleanest control would be to evaluate on networks the LLM could not have seen during training, and were released after its training cutoff or held privately. Synthetic Bayesian networks offer one alternative: procedurally generating fresh graphs whose variables and dependencies are novel by construction and therefore cannot have been memorized. The main difficulty is keeping such domains semantically coherent, since a network over meaningless variables would suppress the semantic signal \name relies on and would test robustness to nonsense rather than to contamination. Designing generators that produce novel yet semantically plausible networks is an important open problem, and we leave a full contamination-robust benchmark to future work.

\section*{Ethical Considerations}

\noindent\textbf{Medical benchmarks are not clinical artifacts.}
Four of our five benchmarks (\textsc{Hepar2}, \textsc{Neuropathic}, \textsc{Diabetes}, \textsc{Munin}) describe medical processes, but Section~\ref{sec-exp} evaluates structure recovery on synthetic tables sampled from a known reference graph.
The learned graphs should not be used for diagnosis, treatment selection, or causal-effect estimation in real populations without independent clinical validation.

\bibliography{custom}

\appendix
\section{Aggregation Approximation Proof}
\label{app:aggregation-proof}

This appendix proves the aggregation guarantee stated in Section~\ref{sec-method-aggregation}.
It compares the aggregated graph with the full-data output \(G_{\mathrm{std}}=\mathcal A(D)\).
The proof is written for the co-occurrence-normalized voting rule used in the main text.
For any learned graph \(G\), let
\[
E_{\mathrm{skel}}(G)=\{\{u,v\}: u \text{ and } v \text{ are adjacent in } G\}.
\]
Throughout this appendix, for an unordered edge \(e\), the shorthand \(e\in G\)
means \(e\in E_{\mathrm{skel}}(G)\). Thus intersections such as
\(G_{\mathrm{std}}\cap\mathcal M_{\mathrm{cert}}\) are skeleton-level edge
intersections.

\subsection{Sampling and Votes}

Let \(\mathcal M=\{\{i,j\}:1\le i<j\le p\}\) be the candidate edge universe.
In round \(b\), \name{} samples rows \(R_b\subseteq[n]\) and variables \(C_b\subseteq[p]\) uniformly without replacement, runs the base learner \(\mathcal A\) on the induced table \(D[R_b,C_b]\), and embeds the learned graph back into the full variable set.
For \(e=\{u,v\}\in\mathcal M\), define the \emph{eligibility indicator}
\[
J_b(e)=\mathbf{1}\{\{u,v\}\subseteq C_b\},
\]
and the conditional vote indicator
\[
Z_b(e)=\mathbf{1}\{e\in E_{\mathrm{skel}}(\widehat G^{(b)})\}.
\]
Note \(Z_b(e)=0\) whenever \(J_b(e)=0\) (an ineligible round cannot include \(e\)).
Let \(c_e=\sum_{b=1}^B J_b(e)\) denote the co-occurrence count.
\name{} uses \emph{co-occurrence voting}:
\[
\widehat f_e
=
\begin{cases}
\dfrac{1}{c_e}\sum_{b=1}^B J_b(e)\,Z_b(e)
& \text{if } c_e>0,\\
0
& \text{if } c_e=0.
\end{cases}
\]
Define the \emph{conditional edge probability}
\[
p_e^{\mathrm{co}}=\Pr[Z_b(e)=1\mid J_b(e)=1].
\]
On the event \(c_e>0\), the variables \(\{Z_b(e)\}_{b:J_b(e)=1}\) are independent Bernoulli\((p_e^{\mathrm{co}})\).
Hence \(\widehat f_e\) is an unbiased estimator of \(p_e^{\mathrm{co}}\) conditional on \(c_e>0\).
Co-occurrence voting removes the mechanical shrinkage by the endpoint-eligibility rate \(\pi_{\mathrm{elig}}=k(k-1)/(p(p-1))\); however, \(p_e^{\mathrm{co}}\) may still depend on \(k\) through the other variables co-sampled in the subproblem.
The concentration argument below accounts for the probability that \(c_e=0\).

\subsection{Certified Edge Universe}

The theorem is stated on a subset of edges whose full-data decision can be localized to a certified step of the full-data greedy search.
Let \(\operatorname{Trace}_{\mathrm{std}}\) denote the instrumented full-data run of \(\mathcal A\), including evaluated but non-selected GES operators.

For an edge \(e=\{X,Y\}\), an operator-level decision certificate is a tuple
\[
d=(e,\mathrm{tag},c,\mathrm{Ctx},a,z),
\]
where \(\mathrm{tag}\in\{\mathrm{Insert},\mathrm{Delete}\}\), \(c\in\{X,Y\}\) is the child node for the local BDeu score, \(\mathrm{Ctx}\) is the local CPDAG context used by the operator, \(a\) records whether the operator is selected or rejected by the greedy decision at that step, and
\[
z=\mathbf{1}\{e\in E_{\mathrm{skel}}(G_{\mathrm{std}})\}
\]
is the final include/exclude verdict of the full-data learner.
The informative variable set for the certificate is
\[
U_d=\{X,Y\}\cup \mathrm{Pa}_c\cup \mathrm{NA}_{c,c'}\cup S,
\]
where \(c'\) is the other endpoint and \(S\) is the insert/delete conditioning set used by the Chickering operator.

A certificate is certifying for \(e\) if its local decision is consistent with the final bit \(z\) and no later selected full-data operator reverses that bit.
The certified universe is
\[
\mathcal M_{\mathrm{cert}} = \{e\in\mathcal M : \exists \text{ certifying decision for }e\}.
\]
For each \(e\in\mathcal M_{\mathrm{cert}}\), fix one canonical certifying decision \(d_e\), set \(U_e=U_{d_e}\), and partition
\[
\begin{aligned}
\mathcal M_{\mathrm{cert}}^+
&=\mathcal M_{\mathrm{cert}}\cap E_{\mathrm{skel}}(G_{\mathrm{std}}),\\
\mathcal M_{\mathrm{cert}}^-
&=\mathcal M_{\mathrm{cert}}\setminus E_{\mathrm{skel}}(G_{\mathrm{std}}).
\end{aligned}
\]
Finally, let \(M_{\mathrm{cert}}=|\mathcal M_{\mathrm{cert}}|\le\binom{p}{2}\).

\subsection{Concentration Proof}

For \(e\in\mathcal M_{\mathrm{cert}}\), define the one-round conditional error probabilities:
\begin{align*}
q_e^{\mathrm{co}+} &:= \Pr[Z_b(e)=0\mid J_b(e)=1],\\
q_e^{\mathrm{co}-} &:= \Pr[Z_b(e)=1\mid J_b(e)=1].
\end{align*}
These are bounded using row-noise and path-stability terms:
\begin{align*}
q_e^{\mathrm{co}+} &\le 1-\frac{\pi_{\mathrm{info}}(e)}{\pi_{\mathrm{elig}}(e)}(1-\chi_e),\\
q_e^{\mathrm{co}-} &\le 1-\frac{\pi_{\mathrm{info}}(e)}{\pi_{\mathrm{elig}}(e)}(1-\chi_e),
\end{align*}
where \(\chi_e=\min\{1,\rho_e+\xi_m(e)\}\) combines path instability and row-subsampling noise.
The conditional edge probability given eligibility is
\[
p_e^{\mathrm{co}} := \Pr[Z_b(e)=1\mid J_b(e)=1] \ge 1-q_e^{\mathrm{co}+}.
\]
For \(e\in\mathcal M_{\mathrm{cert}}^+\), the vote margin is \(a_e^+ = p_e^{\mathrm{co}}-\tau\), and for \(e\in\mathcal M_{\mathrm{cert}}^-\), the vote margin is \(a_e^- = \tau-p_e^{\mathrm{co}}\).
Assume \(a_e^\pm > 0\) for every certified edge.

Conditional on the eligible-round set \(\{b:J_b(e)=1\}\), the variables
\(\{Z_b(e)\}_{b:J_b(e)=1}\) are independent Bernoulli\((p_e^{\mathrm{co}})\).
By Hoeffding's inequality,
\[
\Pr[\text{edge error}\mid c_e=c] \le \exp(-2c(a_e^\pm)^2).
\]
When \(c_e=0\), the convention \(\widehat f_e=0\) ensures the edge is excluded, which is covered by the same marginalization below.
The co-occurrence count \(c_e\sim \mathrm{Binomial}(B,\pi_{\mathrm{elig}})\).
Marginalizing over \(c_e\) using the moment-generating function,
\begin{align*}
&\mathbb{E}[\exp(-2c_e(a_e^\pm)^2)] \\
&= \left(1-\pi_{\mathrm{elig}}+\pi_{\mathrm{elig}}\exp(-2(a_e^\pm)^2)\right)^B\\
&= \left(1 + \pi_{\mathrm{elig}}(\exp(-2(a_e^\pm)^2)-1)\right)^B.
\end{align*}
Therefore,
\begin{equation}
\label{app:eq:binomial-mgf}
\begin{aligned}
\Pr[\text{edge error}]
&\le
\Bigl(1-\pi_{\mathrm{elig}} \\
&\qquad {}\times
\left[1-\exp\left(-2(a_e^\pm)^2\right)\right]
\Bigr)^B .
\end{aligned}
\end{equation}

For the worst-case certified margin \(a_{\min}=\min_e a_e^\pm\), taking a union bound over certified edges,
\begin{equation}
\label{app:eq:certified-union-bound}
\begin{aligned}
&\Pr\!\left[
  \widehat{G}_B(\tau)\cap\mathcal M_{\mathrm{cert}}
  \ne
  G_{\mathrm{std}}\cap\mathcal M_{\mathrm{cert}}
\right] \\
&\qquad\le
M_{\mathrm{cert}}
\left(
  1-\pi_{\mathrm{elig}}\left(1-\exp(-2a_{\min}^2)\right)
\right)^B .
\end{aligned}
\end{equation}
This proves Eq.~\ref{eq:aggregation-approx-bound}.
The per-edge form \(\sum_e \left(1-\pi_{\mathrm{elig}}(1-\exp(-2a_e^2))\right)^B\) is tighter.

\subsection{Aggregation Budget}

From Eq.~\ref{app:eq:certified-union-bound}, to achieve certified error probability \(\delta\) on the certified edge set, it suffices that
\[
M_{\mathrm{cert}}\left(1-\pi_{\mathrm{elig}}(1-\exp(-2a_{\min}^2))\right)^B \le \delta.
\]
Using the standard bound \((1-x)^B\le\exp(-Bx)\), we obtain the sufficient confidence rule
\begin{equation*}
B_{\delta}(a_{\min}) = \left\lceil \frac{\log(M_{\mathrm{cert}}/\delta)}{\pi_{\mathrm{elig}}(1-\exp(-2a_{\min}^2))} \right\rceil.
\end{equation*}
For \(\delta=0.05\) and a pre-specified upper bound \(M_0\) on certified edges,
\begin{equation}
\label{app:eq:b95-target-margin}
B_{95}(a_{\mathrm{target}}) = \left\lceil \frac{\log(20M_0)}{\pi_{\mathrm{elig}}(1-\exp(-2a_{\mathrm{target}}^2))} \right\rceil.
\end{equation}
The realized weakest margin \(a_{\min}\) is not known before the aggregation is run.
For small margins \(a_{\mathrm{target}} \ll 1\), the denominator satisfies \(1-\exp(-2a_{\mathrm{target}}^2) \approx 2a_{\mathrm{target}}^2\), giving the approximate design heuristic
\[
B_{95}(a_{\mathrm{target}}) \approx
\left\lceil
\frac{\log(20M_0)}{2\pi_{\mathrm{elig}}\,a_{\mathrm{target}}^2}
\right\rceil.
\]
If the realized \(a_{\min}\ge a_{\mathrm{target}}\) and \(M_{\mathrm{cert}}\le M_0\), then the design round count \(B_{95}(a_{\mathrm{target}})\) certifies 95\% agreement on the certified edge set.
If either condition fails post-hoc, observed vote margins and certified-edge counts diagnose whether the shortfall comes from unresolved difficult edges or from an underestimated edge count.

\subsection{Sufficient Conditions for Positive Margins}

The concentration argument above only requires positive margins \(a_e^\pm\).
We next give sufficient conditions linking those margins to variable coverage, row noise, and stability of the full-data GES path.

\paragraph{Variable coverage.}
For an edge \(e=\{X,Y\}\), define endpoint eligibility
\[
\begin{aligned}
\mathcal J_e^{(b)}
&=\{\{X,Y\}\subseteq C_b\},\\
\pi_{\mathrm{elig}}(e)
&=\Pr(\mathcal J_e^{(b)})
=\frac{k(k-1)}{p(p-1)}.
\end{aligned}
\]
Let \(\mathcal I_e^{(b)}=\{U_e\subseteq C_b\}\) be informative coverage of the certifying local context.
If \(s_e=|U_e|\), then
\[
\pi_{\mathrm{info}}(e)=\Pr(\mathcal I_e^{(b)})
=
\frac{\binom{p-s_e}{k-s_e}}{\binom{p}{k}},
\]
with value \(0\) when \(k<s_e\).

\paragraph{Path stability.}
Let \(\mathcal P_e^{(b)}\) be the event that, conditional on informative coverage, the subproblem run reaches a matching certifying decision for \(e\) and no later selected operator changes the final bit contrary to \(z_e=\mathbf{1}\{e\in E_{\mathrm{skel}}(G_{\mathrm{std}})\}\).
Define
\[
\rho_e=\Pr[(\mathcal P_e^{(b)})^c\mid \mathcal I_e^{(b)}].
\]
Let \(\xi_m(e)\) denote the row-subsampling probability that the local BDeu decision associated with the certifying operator is reversed despite informative coverage and path stability.
The combined one-round instability is
\[
\chi_e=\min\{1,\rho_e+\xi_m(e)\}.
\]
Then the one-round mismatch probabilities satisfy, for \(e\in\mathcal M_{\mathrm{cert}}^+\),
\[
q_e^+
:=\Pr[Z_b(e)=0]
\le
1-\pi_{\mathrm{info}}(e)(1-\chi_e),
\]
and, for \(e\in\mathcal M_{\mathrm{cert}}^-\),
\[
q_e^-
:=\Pr[Z_b(e)=1]
\le
\pi_{\mathrm{elig}}(e)-\pi_{\mathrm{info}}(e)(1-\chi_e).
\]
Under co-occurrence voting, the relevant probability is \(p_e^{\mathrm{co}}\), the conditional edge probability given eligibility.
The one-round mismatch probability (given \(J_b(e)=1\)) satisfies:
\begin{align*}
q_e^{\mathrm{co}+} &:= \Pr[Z_b(e)=0\mid J_b(e)=1],\\
&\le 1-\frac{\pi_{\mathrm{info}}(e)}{\pi_{\mathrm{elig}}(e)}(1-\chi_e),\\
q_e^{\mathrm{co}-} &:= \Pr[Z_b(e)=1\mid J_b(e)=1],\\
&\le 1-\frac{\pi_{\mathrm{info}}(e)}{\pi_{\mathrm{elig}}(e)}(1-\chi_e).
\end{align*}
Thus, for \(e\in\mathcal M_{\mathrm{cert}}^+\), a sufficient vote margin is
\[
\underline a_e^+ = \frac{\pi_{\mathrm{info}}(e)}{\pi_{\mathrm{elig}}(e)}(1-\chi_e)-\tau,
\]
and, for \(e\in\mathcal M_{\mathrm{cert}}^-\),
\[
\underline a_e^- = \tau-\left[1-\frac{\pi_{\mathrm{info}}(e)}{\pi_{\mathrm{elig}}(e)}(1-\chi_e)\right].
\]
The ratio \(\pi_{\mathrm{info}}/\pi_{\mathrm{elig}}\) is the probability that, given co-sampling of both endpoints, the full certifying context \(U_e\) is also co-sampled.
If every \(\underline a_e^\pm\) is positive, the theorem holds with
\(a_{\min}\) replaced by
\[
\begin{aligned}
\underline a_{\min}
&=
\min\Bigl\{
\min_{e\in\mathcal M_{\mathrm{cert}}^+}\underline a_e^+,\,
\min_{e\in\mathcal M_{\mathrm{cert}}^-}\underline a_e^-
\Bigr\},
\end{aligned}
\]
with empty positive or negative classes omitted.

\subsection{BDeu and Path-Stability Ingredients}

We record the analytic ingredients used to bound \(\xi_m(e)\) and \(\rho_e\).

\paragraph{BDeu score-scale shift.}
Let \(v_{\mathrm{sub}}=k-1\) and \(v_{\mathrm{full}}=p-1\) be the variable universe sizes entering the structure prior for the subproblem and full-data learner.
Under finite state spaces, \(\alpha_{\mathrm{ess}}>0\), and positive local cell mass, define
\[
\begin{aligned}
\Delta_\Phi(m,n,k,p)
&=
\frac{\log(v_{\mathrm{sub}}/\eta_{\mathrm{str}}-1)}{m}
\\[-1mm]
&\quad -
\frac{\log(v_{\mathrm{full}}/\eta_{\mathrm{str}}-1)}{n}
\end{aligned}
\]
Then the deterministic per-row BDeu shift between the subproblem scale and the full-data scale satisfies
\[
\begin{aligned}
\psi_e(m,n,k,p)
&\le
|\Delta_\Phi(m,n,k,p)|\\
&\quad
+C_1\left(\frac{\log m}{m}+\frac{\log n}{n}\right)\\
&\quad
+C_2\left(\frac{1}{m}+\frac{1}{n}\right),
\end{aligned}
\]
where \(C_1\) and \(C_2\) depend on the local table size and BDeu pseudo-counts.
This is the term that prevents a small full-data local margin from being treated as certifiable at a much smaller row/variable scale.

\paragraph{Local BDeu Lipschitz control.}
Let \(\mu_e>0\) be the minimum local joint cell mass for the certifying local table, and let \(r_Y q_Y r_X\) be the number of cells in the local \((Y,X,\mathrm{Pa}_Y)\) table.
On the local probability simplex ball \(\|\nu-\nu^{\mathrm{std}}\|_\infty\le \mu_e/2\) and for \(m\ge 2/\mu_e\), the normalized BDeu local contrast is Lipschitz with modulus
\begin{equation}
\label{app:eq:bdeu-lipschitz}
\begin{aligned}
L_e^{\mathrm{BDeu}}(m,k)
&\le
r_Yq_Yr_X\\
&\quad \times
\Biggl[
2|\log(\mu_e/2)|
\\
&\quad+
\frac{16(1+\alpha_{\mathrm{ess}})}{m\mu_e}
\Biggr].
\end{aligned}
\end{equation}
The key cancellation is that the leading \(\log m\) terms in the paired digamma derivatives cancel; for fixed \(\mu_e\), the limiting modulus is independent of \(m\).

\paragraph{Row-noise bound.}
Let \(\widetilde\gamma_e=[|\bar\Delta_e^{\mathrm{std}}|-\psi_e]_+\) be the effective full-data local margin after subtracting the deterministic score-scale shift, and define
\[
t_e=
\min\left\{
\frac{\mu_e}{2},
\frac{\widetilde\gamma_e}{L_e^{\mathrm{BDeu}}(m,k)}
\right\}.
\]
For a constant \(C_e\) equal to the relevant local cell count,
\[
\xi_m(e)\le 2C_e\exp(-2m t_e^2).
\]

\paragraph{Path-stability sufficient condition.}
Let \(g_*^{\mathrm{path}}>0\) be the minimum greedy decision gap along the full-data GES path, \(\psi_*^{\mathrm{path}}\) the worst score-scale shift, \(\widetilde g_*^{\mathrm{path}}=[g_*^{\mathrm{path}}-2\psi_*^{\mathrm{path}}]_+\), \(L_*^{\mathrm{path}}\) the worst local Lipschitz modulus, and \(\Sigma_A^{\mathrm{path}}\) the total number of candidate operators evaluated along the full-data path.
With
\[
t_*^{\mathrm{path}}
=
\min\left\{
\frac{\mu_*^{\mathrm{path}}}{2},
\frac{\widetilde g_*^{\mathrm{path}}}{2L_*^{\mathrm{path}}(m,k)}
\right\},
\]
the path-instability parameter satisfies
\[
\begin{aligned}
\rho_e
&\le
\rho_e^{\mathrm{col}}
+4\Sigma_A^{\mathrm{path}}C_*^{\mathrm{path}}\\
&\quad \times
\exp[-2m(t_*^{\mathrm{path}})^2].
\end{aligned}
\]
Here \(\rho_e^{\mathrm{col}}\) is the full-trace variable-omission floor: even with no row noise, a fixed variable subsample can omit variables needed to shadow the full-data greedy path.

\subsection{Variable-Subsampling Regimes}

\paragraph{Fixed (k<p).} Co-occurrence voting eliminates the \emph{denominator bias} that all-round voting introduces at small \(k/p\).
Under all-round voting, \(\mathbb{E}[\widehat f_e^{\mathrm{ar}}]=\pi_{\mathrm{elig}}\cdot p_e^{\mathrm{co}}\), so the threshold \(\tau\) must be rescaled by \(\pi_{\mathrm{elig}}\) to maintain the same inclusion criterion --- a per-dataset correction that is impractical to specify ahead of time.
Co-occurrence voting targets \(p_e^{\mathrm{co}}\) directly, making \(\tau\) dataset-agnostic.

However, co-occurrence voting does not eliminate the \emph{informative-coverage floor} \(\rho_e^{\mathrm{col}}\).
Even when both endpoints are co-sampled, the certifying local context \(U_e\) (which may include parents or conditioning variables beyond \(\{u,v\}\)) can still be absent from \(C_b\).
Certified agreement in the co-occurrence regime therefore requires only that the sufficient margins \(\underline a_e^\pm\) remain positive after accounting for the informative-coverage shortfall --- not that the denominator bias vanishes (it already has).

\paragraph{Full-information limit.} As \(k\to p\), \(m\to n\), and \(B\to\infty\), we have \(\pi_{\mathrm{elig}}(e)\to 1\), \(\pi_{\mathrm{info}}(e)\to 1\), \(\rho_e^{\mathrm{col}}\to 0\), and the row-noise terms vanish, giving
\[
\Pr\!\left[
\widehat G_B(\tau)\cap\mathcal M_{\mathrm{cert}}
=
G_{\mathrm{std}}\cap\mathcal M_{\mathrm{cert}}
\right]\to 1.
\]
The full-information limit is the same under all-round and co-occurrence voting; before that limit, co-occurrence voting keeps \(\tau\) interpretable without requiring \(k \approx p\).

\subsection{Diagnostics}
\label{app:agg-diagnostics}
For each run, assess the theorem's dataset-dependent conditions using the following diagnostics:
\begin{itemize}
\item the chosen \(B\), \(\tau\), \(m\), \(k\), \(a_{\mathrm{target}}\), and
      \(M_0\);
\item the observed vote-margin distribution
      \(|\widehat f_e-\tau|\) on certified edges, including the minimum or
      low quantiles;
\item \(M_{\mathrm{cert}}\), or the conservative upper bound used if the
      certified trace is not instrumented;
\item endpoint and informative coverage counts for certified edges;
\item when available, path-instability summaries such as
      \(\widehat\rho_e\) or separate variable-floor and row-path estimates.
\end{itemize}
Do not use these diagnostics to tune \(B\) after test-set evaluation.

\subsection{Outside the Certified Universe}

The theorem makes no claim about \(\mathcal M\setminus\mathcal M_{\mathrm{cert}}\).
Such edges lack a localized full-data certifying decision under the instrumented GES trace.
Extending the guarantee to all non-edges would require additional assumptions about the full-data learner or the data-generating graph.
\name{}'s main theorem therefore remains scoped to certified agreement with \(G_{\mathrm{std}}\).

Every edge present in \(G_{\mathrm{std}}\) has at least one selected Insert in the full-data trace that is not later reversed, so it belongs to \(\mathcal M_{\mathrm{cert}}^+\).
The uncovered set \(\mathcal M\setminus\mathcal M_{\mathrm{cert}}\) therefore consists of full-data non-edges that the trace does not certify locally.

If \(G^*\) denotes a true causal graph, write
\(S_B=E_{\mathrm{skel}}(\widehat G_B(\tau))\),
\(S_{\mathrm{std}}=E_{\mathrm{skel}}(G_{\mathrm{std}})\), and
\(S^*=E_{\mathrm{skel}}(G^*)\). Then
\[
\begin{aligned}
|S_B\triangle S^*|
&\le
|S_B\triangle S_{\mathrm{std}}|
+
|S_{\mathrm{std}}\triangle S^*|.
\end{aligned}
\]
This appendix controls the first term on \(\mathcal M_{\mathrm{cert}}\).
Controlling the second term requires separate consistency assumptions for the full-data BN learner and is outside the learner-relative guarantee.

\section{Implementation Details}
\label{app:implementation}

This appendix records the implementations behind every method evaluated in
Section~\ref{sec-exp}, including the statistical and LLM-driven baselines as
well as each component of \name.

\subsection{Statistical Base Learners}
\label{app:implementation-stat}

\noindent\textbf{GES:}
\name's base learner and the \textsc{GES} baseline both use the GES
implementation from the \texttt{causal-learn} library using BDeu as the scoring method.
Inside \name, each subsampling round invokes a fresh GES call on the sampled
subtable while the full-data \textsc{GES} baseline runs the same procedure on
the entire synthetic data table at once.

\noindent\textbf{Out-of-memory investigation with \textsc{GES} on \textsc{Hepar2}:}
The full-data \textsc{GES} baseline fails with an out-of-memory error on
\textsc{Hepar2}. This is not caused by an unbounded parent search, as \textsc{GES}
runs under the same per-node in-degree cap $\kappa$ as \name{}'s internal base
learner (Table~\ref{tab:hyperparameters}). The cause is instead specific to the
\texttt{causal-learn} \textsc{GES+BDeu} implementation. It memoizes local family scores, and \textsc{BDeu} scoring allocates contingency tables whose size grows multiplicatively with the cardinalities of the child and parent variables.
\textsc{Hepar2} contains multi-state categorical variables with up to seven states, so candidate families can induce large intermediate count tables. Eventually the combined footprint exceeds available memory on our hardware.
This failure mode is specific to the evaluated implementation and score. While alternative implementations or scores such as \textsc{BIC} would possibly have a smaller memory footprint on the same benchmark, we keep \textsc{BDeu} fixed across the \textsc{GES}-based methods so that the comparison isolates the effect of subsample aggregation rather than a change of scoring criterion.

\noindent\textbf{PC:}
The \textsc{PC} baseline uses the constraint-based PC implementation in the
\texttt{pgmpy} library with all default parameters. The relevant defaults
are a hard depth cap on the conditioning-set size
(\texttt{max\_cond\_vars}\(=5\)), a parallel CI-test variant
(\texttt{variant=parallel}), and a significance level of
\texttt{\(0.01\)}. The depth cap in particular is
vital on the larger networks: without it, \textsc{PC}'s skeleton
discovery enumerates an exponentially growing number of subsets of each
node's neighbours at each depth, and the runtime quickly becomes
intractable on networks with hub vertices.

\noindent\textbf{Parameter fitting:}
After structure learning, every method that produces conditional probability
tables fits parameters by maximum
likelihood with a uniform Dirichlet smoothing prior, using the standard
\texttt{pgmpy} estimator. \name{} fits a separate CPT to each node, using only the variables in that node's Markov blanket. This allows \name{} to easily scale to arbitrarily large networks, while still supporting prediction.
\textsc{GES}, \textsc{PC}, and \name{} all add any
missing nodes to the completed DAG as isolated nodes before parameter fitting.

\subsection{LLM Baselines}
\label{app:implementation-llmbaselines}

\noindent\textbf{PromptBN:}
We reimplement \textsc{PromptBN} \citep{zhang2026bayesian} as a single
LLM call that is given the variable list together with the variable
descriptions of Section~\ref{app:datasets-vardesc} and is asked to return the
full directed edge set in one shot. If the returned edge set is invalid, the LLM is prompted to retry up to five times. The model is given a per-response budget of \(16384\) tokens.

\noindent\textbf{bfsBN:}
\textsc{bfsBN} \citep{jiralerspong2024efficient} is reimplemented as a
breadth-first, multi-turn conversation in which the LLM is asked, at each
turn, for the parents of the next variable to be expanded. The model is given
a per-response budget of \(8192\) tokens. Because every turn is retained in
the conversation, \textsc{bfsBN}'s context usage grows with the number of
variables explored, making it sensitive to both network size and per-turn
response length.

The original \textsc{bfsBN} study reports success on \textsc{Neuropathic}
with GPT-4. In our setting, \textsc{bfsBN} completes \textsc{Neuropathic} with
DeepSeek but exhausts the context window with GPT-5.4 and
Gemini. We read this as a context-budget effect rather than a limitation of
the algorithm: the reasoning models we use likely emit substantially longer per-turn
responses than GPT-4, so the conversation history grows faster and, on a
network as large as \textsc{Neuropathic}, can overrun the available window.

\noindent\textbf{LLM-CD:}
\label{app:implementation-llmcd}
We reimplement \textsc{LLM-CD} \citep{du2025llmcd} from the released
GitHub repo, which inserts LLM guidance into a typical PC algorithm. The pipeline runs \texttt{causal-learn}'s
\texttt{SkeletonDiscovery} with LLM consultation on borderline CI tests,
followed by an LLM edge-orientation pass on the resulting skeleton.
The original implementation had an optional outside refine-loop to 
improve the classification ability of the model on a single node. However, as
this paper focuses only on whole-graph structure learning and not downstream
single-node predictive capabilities, this external loop is omitted from this implementation.
The runtime-governing parameters match the upstream values:
\(\alpha = 0.05\), chi-square CI test, LLM borderline threshold
\(0.001\), no bound on conditioning-set size, and a single-threaded
skeleton phase. The model is given a per-response budget of \(4096\) tokens.

\textbf{Timeout failure investigation with \textsc{LLM-CD} on \textsc{Neuropathic}:} Our \textsc{LLM-CD} reimplementation
completes \textsc{Hepar2} in 30 hours but does not finish running on \textsc{Neuropathic}
within a 48-hour wall-clock budget, despite the PC baseline finishing on that dataset in time. The non-termination follows from the published algorithm's
choice of skeleton search. \textsc{LLM-CD} inherits
\texttt{causal-learn}'s \texttt{SkeletonDiscovery} with no bound on the
conditioning-set size, so at the high PC
depths reached on hub vertices the per-node CI-test count grows
combinatorially. \textsc{LLM-CD} \citep{du2025llmcd} was evaluated on
networks up to 51 nodes, so \textsc{Neuropathic} (\(p=222\)) sits outside
the regime in which it was demonstrated.

However, this is not inconsistent with our \textsc{PC} baseline completing
\textsc{Neuropathic} in \(35\) hours.
\texttt{pgmpy}'s \textsc{PC} defaults that we use (a hard depth cap, parallel CI dispatch, and a tighter
\(\alpha = 0.01\)) together prevent the deep
enumeration that consumes \textsc{LLM-CD}'s wall-clock. Capping depth would likely significantly reduce required runtime,
but both are deviations from the algorithm as published, so we report
\textsc{LLM-CD} as it stands and mark \textsc{Neuropathic} as not completed.

\subsection{The \name{} Pipeline}
\label{app:implementation-riolu}

Algorithm~\ref{alg:riolu} gives the full \name{} pipeline. The four
LLM-augmentation hooks named in Section~\ref{subsec-llmbag} appear as labeled
subroutines.
Each subroutine is independently enabled by a
configuration flag --- with all flags disabled the pipeline reduces to the
\namenollm{} variant from Section~\ref{subsec-baselines}.

\begin{algorithm}[ht]
\scriptsize
\caption{\name{} pipeline}
\label{alg:riolu}
\begin{algorithmic}[1]
\Require Table \(D\) with \(p\) variables, subsampling budget \(B\), subsample sizes \((m,k)\), vote threshold \(\tau\), max parents \(\kappa\)
\If{column grouping enabled}
  \State \(\mathcal G_{\mathrm{LLM}}\gets\Call{LLM-ColumnGrouping}{D}\)
\EndIf
\State Initialize directed-edge tallies \(\widehat n_{u\to v}\gets 0\) and co-occurrence counts \(c_{u,v}\gets 0\)
\For{\(b=1,\ldots,B\)}
  \State Sample rows \(R_b\) of size \(m\) without replacement
  \State Sample columns \(C_b\) of size \(k\); upweight pairs in \(\mathcal G_{\mathrm{LLM}}\) if enabled
  \State \(\widehat G^{(b)}\gets\textsc{GES}\!\left(D[R_b,C_b];\,S_{\mathrm{BDeu}}\right)\)
  \For{edge \(u\to v\in\widehat G^{(b)}\)} \(\widehat n_{u\to v}\gets\widehat n_{u\to v}+1\) \EndFor
  \For{pair \(\{u,v\}\subseteq C_b\)} \(c_{u,v}\gets c_{u,v}+1\) \EndFor
\EndFor
\If{adaptive sampling enabled}
  \State \(\widehat n,c\gets\Call{LLM-AdaptiveSampling}{D,\widehat n,c}\) 
\EndIf
\State \(\widehat f_{u\to v}\gets\widehat n_{u\to v}/c_{u,v}\) for every candidate edge with \(c_{u,v}>0\)
\State \(\mathcal E_+\gets\{u\to v:\widehat f_{u\to v}>0\}\) and \(\mathcal E_\tau\gets\{u\to v:\widehat f_{u\to v}\ge\tau\}\)
\State For each pair \(\{u,v\}\) with both \(u\to v\) and \(v\to u\) in \(\mathcal E_\tau\), keep only the direction with the larger \(\widehat n\)
\For{each node \(v\)}
  \If{parent ordering enabled}
    \State \(L_v\gets\Call{LLM-ParentOrdering}{v,\mathcal E_+}\)
  \Else
    \State \(L_v\gets\) candidate parents of \(v\) in \(\mathcal E_\tau\), ranked by Pearson correlation with \(v\) in \(D\)
  \EndIf
\EndFor
\State \(G\gets\) empty DAG on the \(p\) variables
\For{each node \(v\) in descending order of \(\sum_{w} \widehat n_{v\to w}\)}
  \For{each candidate parent \(u\in L_v\) with \(|\mathrm{Pa}_G(v)|<\kappa\)}
    \If{adding \(u\to v\) would create no cycle}
      \State add \(u\to v\) to \(G\)
    \ElsIf{cycle arbitration enabled}
      \State \(G\gets\Call{LLM-CycleArbitration}{G,u,v}\)
    \EndIf
  \EndFor
\EndFor
\State \Return \(G\)
\end{algorithmic}
\end{algorithm}

\subsection{Column Grouping}
\label{app:implementation-grouping}

Column grouping identifies, for every variable, the other variables that are
likely to share a direct causal relationship with it. We query the LLM once
per variable, then the union of all returned sets is combined into a pair-level
\emph{related-pair} relation. This biases column subsampling so that
causally adjacent variables co-occur more often than under uniform sampling.
Concretely, during column subsampling each candidate is reweighted by a
factor \(\alpha\) (Table~\ref{tab:hyperparameters}) whenever it is related
to at least one already-picked column. 

\subsection{Adaptive Sampling}
\label{app:implementation-adaptive}

After the initial \(B\) subsampling rounds, adaptive sampling measures the
Bernoulli entropy of every surviving edge's vote frequency and flags as
uncertain any node with at least one incoming edge whose entropy
exceeds the threshold \(H_{\mathrm{unc}}\) (in bits) and whose endpoints
co-occur in at least \(c_{\min}\) samples. For each flagged node, the LLM
is shown the uncertain candidate parents with their observed frequencies and
asked which other variables should be co-sampled to resolve the ambiguity.
The LLM-suggested companions are then force-included in \(t\) additional
subsampling rounds per flagged node, with the remaining column slots drawn by
the same procedure as the initial pool. Each targeted round runs the same
GES call as the initial sampling. The resulting edges and co-sampled pairs
are merged into the running tallies before \name{} proceeds.
Algorithm~\ref{alg:adaptive} summarizes the procedure.

\begin{algorithm}[ht]
\scriptsize
\caption{\textsc{LLM-AdaptiveSampling}}
\label{alg:adaptive}
\begin{algorithmic}[1]
\Require Data \(D\), tallies \(\widehat n\), co-counts \(c\), entropy threshold \(H_{\mathrm{unc}}\), co-occurrence floor \(c_{\min}\), samples per node \(t\), subsample sizes \((m,k)\)
\State \(\mathcal U\gets\emptyset\) \Comment{uncertain nodes}
\For{each candidate edge \(u\to v\) with \(c_{u,v}\ge c_{\min}\)}
  \State \(\widehat f\gets\widehat n_{u\to v}/c_{u,v}\)
  \If{\(H_2(\widehat f)\ge H_{\mathrm{unc}}\)} \(\mathcal U\gets\mathcal U\cup\{v\}\) \EndIf
\EndFor
\For{each \(v\in\mathcal U\)}
  \State \(S_v\gets\) LLM-suggested companions for \(v\)
  \For{\(j=1,\ldots,t\)}
    \State Sample rows \(R\) of size \(m\); sample columns \(C\) of size \(k\) with \(\{v\}\cup S_v\) force-included
    \State \(\widehat G\gets\textsc{GES}(D[R,C];\,S_{\mathrm{BDeu}})\)
    \For{edge \(u'\to v'\in\widehat G\)} \(\widehat n_{u'\to v'}\gets\widehat n_{u'\to v'}+1\) \EndFor
    \For{pair \(\{u',v'\}\subseteq C\)} \(c_{u',v'}\gets c_{u',v'}+1\) \EndFor
  \EndFor
\EndFor
\State \Return \(\widehat n,c\)
\end{algorithmic}
\end{algorithm}

\subsection{Parent Ordering}
\label{app:implementation-parent}

Parent ordering replaces the default correlation-ranked candidate parent
list with an LLM-produced ranking. For each child \(v\), the LLM is
shown \(v\), every candidate parent with a positive subsampling vote
frequency (i.e. every \(u\) with \(\widehat f_{u\to v}>0\) prior to
applying the cutoff), and the subsampling support of each candidate. The
LLM is asked to return the candidates in order of causal relevance.

The vote-frequency cutoff \(\tau\) is applied differently in the
LLM-augmented and unaugmented paths. Without parent ordering, the
greedy assembler receives only the candidates in \(\mathcal E_\tau\).
With parent ordering enabled, candidates the LLM endorses are kept
regardless of their subsampling support, and candidates the LLM omits
are tail-appended in descending \(\widehat f\) and kept only if
\(\widehat f_{u\to v}\ge\tau\). The motivation is that LLM endorsement
should bypass the cutoff, with subsampling support as the floor for everything
else, so neither signal can silently delete a candidate the other
strongly supports. The resulting ordering is consumed directly by the
greedy DAG assembly, with the LLM's top-ranked parents tried first
under the in-degree cap \(\kappa\).

When adaptive sampling is also enabled, candidates that adaptive
sampling targeted are additionally annotated with a pre-/post-
adaptive-sampling trajectory, so the LLM can see whether the targeted
resampling resolved the ambiguity in favor of or against each
candidate.

\subsection{Cycle Arbitration}
\label{app:implementation-cycle}

The greedy assembly of Section~\ref{sec-method-aggregation} adds candidate
edges in vote-rank order and skips any edge that would close a
cycle. Instead, if cycle arbitration is enabled, when adding
\(P\to C\) would close one or more cycles, we enumerate the directed paths
\(C\leadsto P\) currently in \(G\) and produce the union of their edges. The
LLM is shown this edge set and asked for a hitting set whose removal breaks
every offending path. The requested edges are then removed and \(P\to C\) is
added. The proposed removals are verified after the fact: if the LLM
declines (empty hitting set), or if removing its chosen edges does not
actually break every offending path, the removals are reverted and the
candidate edge is dropped as in the baseline. Path enumeration is capped
at \(\Pi_{\max}\) paths---if the cap is exceeded the augmentation falls back
to the baseline reject behavior, since the prompt and the model's ability to
construct a valid hitting set both degrade beyond that point.

\subsection{Fallbacks}

When an LLM response cannot be parsed into the expected schema
or violates a graph-level invariant (DAG-ness, node membership, or
isolation), each augmentation falls back to its statistical counterpart for
that step rather than producing an invalid graph. Exact decoding parameters
and per-augmentation hyperparameters are listed in
Table~\ref{tab:hyperparameters}.

\begin{table}[ht]
\centering
\scriptsize
\begin{tabular}{@{}lll@{}}
\toprule
\textbf{Symbol} & \textbf{Description} & \textbf{Value} \\
\midrule
\multicolumn{3}{@{}l}{\emph{Aggregation}} \\
\(B\) & subsampling rounds & 1000 \\
\(m\) & rows per subsample & 10000 \\
\(k\) & columns per subsample & 15 \\
\(\tau\) & vote-frequency threshold & 0.5 \\
\midrule
\multicolumn{3}{@{}l}{\emph{LLM augmentations}} \\
\(\alpha\) & column upweight factor & 5.0 \\
\(H_{\mathrm{unc}}\) & uncertainty entropy threshold (bits) & 0.7 \\
\(c_{\min}\) & min co-occurrence for uncertainty & 5 \\
\(t\) & targeted samples per uncertain node & 10 \\
\(\Pi_{\max}\) & cycle path enumeration cap & 10 \\
\midrule
\multicolumn{3}{@{}l}{\emph{LLM decoding}} \\
& reasoning effort & medium \\
& max completion tokens & 4096 \\
& temperature & default \\
\bottomrule
\end{tabular}
\caption{Hyperparameters used in the main experiments.}
\label{tab:hyperparameters}
\end{table}

\section{Subsample-Aggregation Ablation: Vote-Threshold Convergence and Sensitivity}
\label{app:tau_ablation}

\begin{figure*}[htb]
  \centering
  \includegraphics[width=0.99\textwidth]{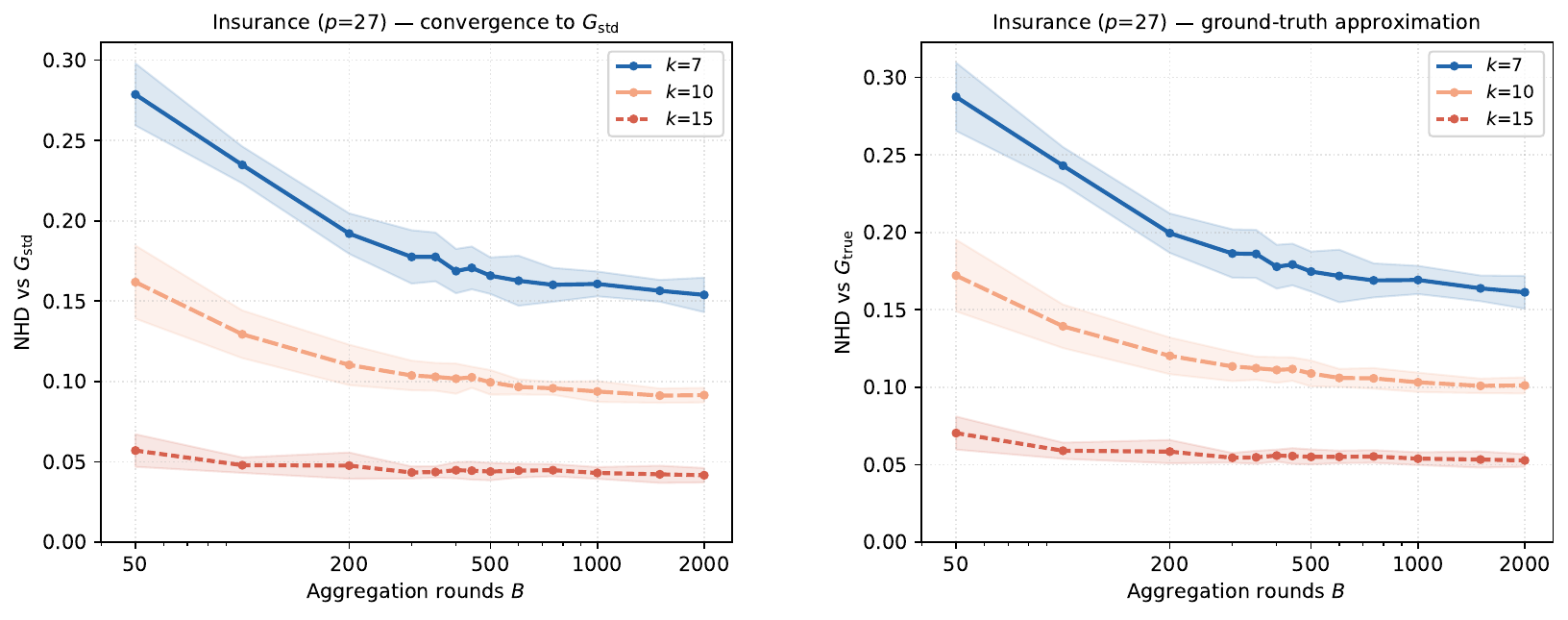}
  \caption{%
    Convergence of subsample-aggregated \textsc{GES+BDeu} with $B$ for
    Insurance ($p$=27) at $k\in\{7,10,15\}$, evaluated at the fixed threshold
    \(\tau=0.5\) used in the main experiments.
    Left: NHD vs.\ $G_{\mathrm{std}}$ (convergence to standard learner).
    Right: NHD vs.\ $G_{\mathrm{true}}$ (ground-truth approximation).
    Shaded bands: $\pm$1 std over 10 seeds.%
  }
  \label{fig:convergence_by_k}
\end{figure*}

\begin{figure*}[htb]
  \centering
  \includegraphics[width=0.99\textwidth]{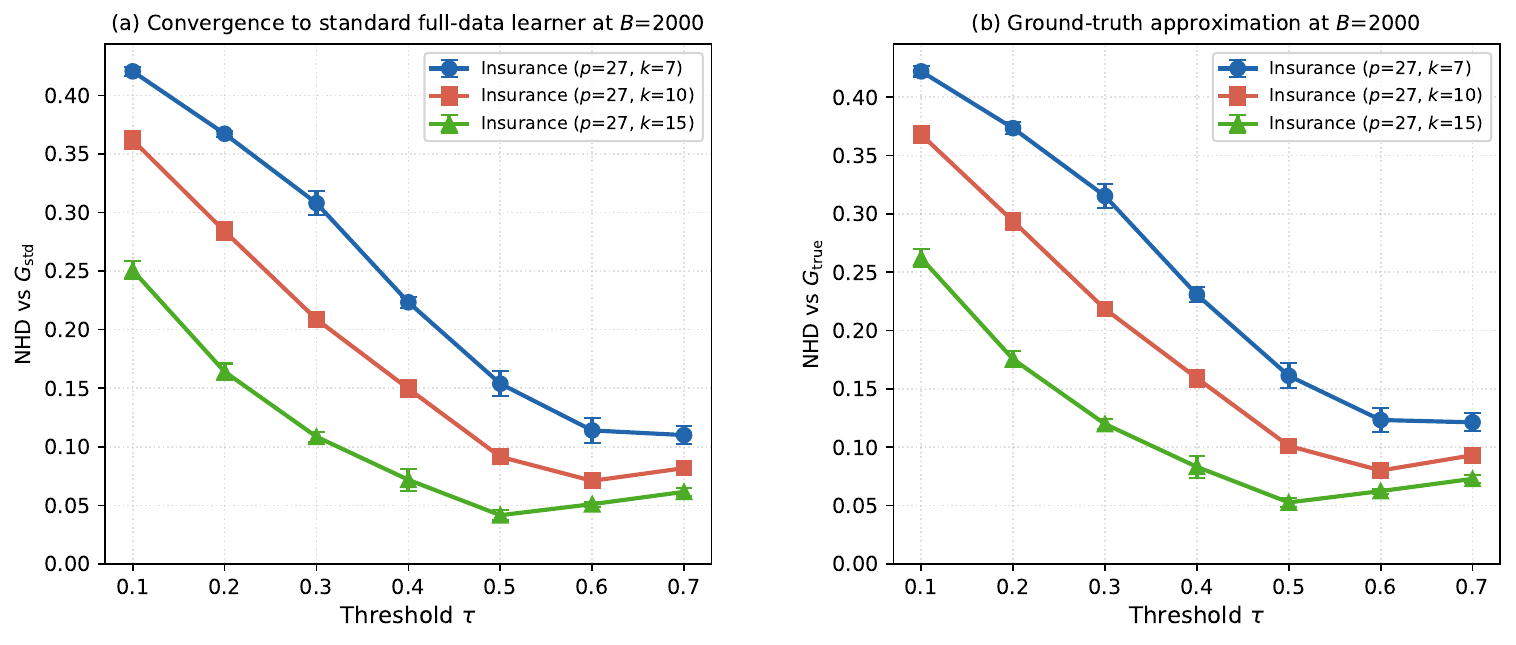}
  \caption{%
    $\tau$ sensitivity at $B=2000$ on Insurance ($p$=27).
    NHD vs.\ $G_{\mathrm{std}}$ (left) and vs.\ $G_{\mathrm{true}}$ (right)
    as a function of $\tau$ for $k\in\{7,10,15\}$.
    Error bars: $\pm$1 std over 10 seeds.
    The optimal $\tau$ decreases from 0.70 to 0.50 as $k$ increases.%
  }
  \label{fig:tau_sensitivity}
\end{figure*}

The aggregation bound in Eq.~\ref{eq:aggregation-approx-bound} predicts that,
for certified edge decisions with positive margins, disagreement with the
full-data reference learner decreases as the number of subsampling rounds grows.
We examine this finite-sample behavior on \textsc{Insurance}. Since the main experiments fix \(\tau=0.5\) before
evaluation, the ablation separates convergence under this threshold from the
sensitivity of the aggregation rule to nearby thresholds.

We use the co-occurrence-normalized voting rule from
Section~\ref{sec-method-aggregation}. For a candidate edge \(u\to v\), the vote
frequency is
\(\widehat f^{\mathrm{co}}_{u\to v}=\hat n_{u\to v}/c_{u,v}\), where \(c_{u,v}\)
counts rounds in which both endpoints were sampled. This normalization removes
the mechanical endpoint-eligibility factor
\(\pi_{\rm elig}=k(k{-}1)/[p(p{-}1)]\) from the vote frequency. The experiments below measure finite-sample convergence and threshold sensitivity on this benchmark; they do not separately verify the theorem's regularity or path-stability conditions.

We sweep three column sizes \(k \in \{7,10,15\}\), thresholds
\(\tau \in \{0.10,0.20,\ldots,0.70\}\), and round counts \(B\) from 50 to 2000.
The metric is Normalized Hamming Distance (NHD) between
\(\widehat G^{\mathrm{co}}_B(\tau)\) and two references: the full-data GES
output \(G_{\mathrm{std}}\) and the ground-truth network \(G_{\mathrm{true}}\).
Results are averaged over 10 independent seeds.

\paragraph{Convergence with $B$.}
Figure~\ref{fig:convergence_by_k} shows NHD vs.\ $G_{\mathrm{std}}$ as $B$
increases under the pre-specified threshold \(\tau=0.5\).
The curves decrease and stabilize as \(B\) grows. Larger \(k\) values produce
lower final NHD, reflecting higher effective co-observation counts. At the
main column budget \(k=15\), the fixed threshold \(\tau=0.5\) gives
\(0.042\pm0.004\) NHD to \(G_{\mathrm{std}}\) and \(0.053\pm0.004\) NHD to
\(G_{\mathrm{true}}\) at \(B=2000\).

\paragraph{Threshold sensitivity.}
Figure~\ref{fig:tau_sensitivity} shows NHD at $B=2000$ as a function of $\tau$.
The best threshold decreases with \(k\):
\(\tau^*=0.70\) for \(k=7\), \(0.60\) for \(k=10\), and \(0.50\) for \(k=15\).
Within each \(k\), nearby thresholds give similar results: the NHD gap between
the best and second-best threshold is at most 0.008. In this setting, modest
misspecification of \(\tau\) therefore has limited effect. When
\(\pi_{\rm elig}\) is small, conservative thresholds suppress noisy edges; when
the column budget is larger, moderate thresholds preserve more intermediate
edges.

For a new dataset, we recommend choosing the largest column budget \(k\) permitted by memory and runtime, since larger subproblems preserve more variable context and increase endpoint co-observation. With the \(k=15\) setting in Table ~\ref{tab:hyperparameters}, $\tau=0.5$ is our recommended initial value. This configuration was used without dataset-specific threshold tuning across all five benchmarks in Table \ref{tab:main-results} and Table \ref{tab:llm-ablation}. When resource constraints require smaller column budgets, the \textsc{Insurance} sweep in Figure \ref{fig:tau_sensitivity} suggests tested starting points of $\tau=0.6$ for \(k=10\) and $\tau=0.7$ for \(k=7\). 
This implies that when each subsample size is smaller, we need larger minimum co-occurrence vote frequency. Nearby thresholds performed similarly: the NHD difference between the best and second-best value was at most $0.008$ for every tested $k$.

\section{Cost Model and Derivations}
\label{app:cost-derivations}

Table~\ref{tab:baseline-comparison} summarizes the practical
structure-search cost (excluding CPT fitting) and the number of LLM API
calls for each method. The $B$ and $t|\mathcal{U}|$ subsample rounds in
\name{} and \namenollm{} are mutually independent and parallelizable.

The per-call structure-search cost under the causal-learn max-parent
cap $\kappa$ is $T_{\mathrm{GES}}(a,b) = O((a + r^{\kappa+1})\,b^4\,2^\kappa)$,
with $a$ rows, $b$ variables, and maximum variable cardinality $r$.
Under maximum graph degree $d$ for the PC skeleton phase,
$T_{\mathrm{PC}}(n,p,d) = O((n + r^{d+2})\,p^2\,2^d)$. Bounded-state
terms ($r$ constant) simplify these to $O(ab^4 2^\kappa)$ and
$O(np^2 2^d)$ respectively.

The \emph{LLM calls} column counts API requests, not token volume. For
\textsc{LLM-CD}, following Du et al.'s discovery-stage accounting, the
count excludes the extra repeated querying used only for uncertainty
analysis. $C$ denotes cycle-arbitration events during graph assembly, 
with $C = O(p^2)$ worst-case. $I$ is the number of
\textsc{LLM-CD} outer iterations, and the subsample dimensions satisfy
$k \ll p$ and $m \leq n$.

\section{Hardware and Cost}
\label{app:hardware}

\noindent\textbf{Compute.}
All baseline runs were executed on a single AWS EC2 instance with 16 vCPUs and 128~GiB RAM. The structure learning uses CPU only, and no GPU is required, and the LLM augmentations dispatch to remote API.

\noindent\textbf{Time and token costs.}
Table~\ref{tab:cost} reports wall-clock time and total LLM token usage for
each method on each benchmark. All methods are run with a per-trial time limit of 48 hours.

\begin{table}[ht]
\centering
\scriptsize
\setlength{\tabcolsep}{3pt}%
\begin{tabular}{@{}llrrr@{}}
\toprule
\textbf{Dataset} & \textbf{Method} & \textbf{Time (min)} & \textbf{Input Tokens} & \textbf{Completion Tokens} \\
\midrule
Insurance & GES & 1.67 & -- & -- \\
 & PC & 3.10 & -- & -- \\
 & PromptBN & 1.72 & 2.9K & 8.6K \\
 & bfsBN & 11.8 & 199K & 36K \\
 & LLM-CD & 12.1 & 24K & 25K \\
 & \namenollm{} & 17.1 & -- & -- \\
 & \name{} & 24.8 & 191K & 140K \\
\midrule
Hepar2 & PC & 48.9 & -- & -- \\
 & bfsBN & 10.8 & 2.0M & 33K \\
 & LLM-CD & 1777 & 92K & 86K \\
 & \namenollm{} & 7.67 & -- & -- \\
 & \name{} & 20.3 & 757K & 308K \\
\midrule
Neuropathic & PC & 2096 & -- & -- \\
 & \namenollm{} & 9.65 & -- & -- \\
 & \name{} & 73.3 & 5.3M & 1.2M \\
\midrule
Diabetes & PromptBN & 8.47 & 79K & 8.6K \\
 & \namenollm{} & 66.6 & -- & -- \\
 & \name{} & 480 & 26M & 2.1M \\
\midrule
Munin & \namenollm{} & 38.4 & -- & -- \\
 & \name{} & 160 & 71M & 3.6M \\
\bottomrule
\end{tabular}
\caption{Wall-clock time and LLM token usage per method per benchmark, for methods that were able to successfully run. LLM-powered methods (PromptBN, bfsBN, \name{}) use GPT-5.4. Non-LLM methods show \texttt{--} for token columns.}
\label{tab:cost}
\end{table}

\section{Dataset Preparation}
\label{app:datasets}

\subsection{Synthetic training data}
\label{app:datasets-synth}

For each benchmark we generate a synthetic training table by forward
sampling \(n\) joint configurations from the ground-truth network. The
sampled table is the only input that the structure learners ever see. The
ground-truth graph is held out and used only for evaluation against the
metrics of Section~\ref{subsec-metrics}. Per-dataset sample counts \(n\) and
the corresponding row/column subsample budgets used during aggregation are
listed in Table~\ref{tab:hyperparameters} and the per-dataset configuration
files released with the code.

\subsection{Variable descriptions}
\label{app:datasets-vardesc}

Every LLM-driven method in this paper consumes a per-network
\emph{variable description} file: a mapping from each variable name in the
network to a short natural-language description of the quantity it
represents and the meaning of its discrete states. Variable descriptions are
authored once per dataset and reused across every LLM augmentation and every
LLM baseline (\textsc{PromptBN}, \textsc{bfsBN}, \textsc{LLM-CD}), so that comparisons across
methods are not confounded by differences in the domain context provided to
the model.

\section{Excluded Augmentation: Graph Refinement}
\label{app:graph-refinement}

A natural extension of the augmentations in
Section~\ref{subsec-llmbag} is a graph refinement step that runs
after greedy assembly and lets the LLM edit the final DAG by adding or
deleting edges, following prior iterative-editing designs
\citep{takayama2025integratinglargelanguagemodels,ban2023iterative}.
This augmentation was excluded from \name{} because every variant 
degraded Edge \(F_1\)  on top of the rest of the pipeline.

The four augmentations we retain each act either on the sampling
distribution ({adaptive sampling}, {column grouping}) or on a
locally bounded decision during graph assembly ({parent ordering}
on one child, {cycle arbitration} on one proposed edge). Graph refinement is the only candidate augmentation whose scope
is the entire assembled graph and whose authority is unilateral
edit, meaning it can add or delete any edge irrespective of upstream
evidence.

By construction, refinement runs last and consumes evidence the rest
of the pipeline has already aggregated such as subsampling support, parent-
ordering endorsement, cycle-arbitration outcomes. When it disagrees
with an upstream decision, it wins by being last. That structural
asymmetry persists regardless of what information the prompt is given
and how narrowly the action space is scoped.

We tested three variants of graph refinement:

\begin{compactenum}
\item \textsc{Vanilla refinement:} the iterative add/delete/terminate
loop described in the implementations cited above.
\item \textsc{Refinement with subsampling support:} every in-graph
edge is annotated with its subsampling support in the prompt, and a
``near-miss'' block annotates edges that were endorsed upstream but
did not fit into the greedy aggregation.
\item \textsc{Narrowly scoped refinement:} targeted refinement specifically
in areas that are likely blind spots for the existing algorithm: confounding
variables and orphan nodes. In two separate passes, the LLM was asked to delete
edges from confounding variables, and to connect orphan nodes to the main graph.
\end{compactenum}

Table~\ref{tab:refinement-deltas} reports the \(F_1\)  delta of each variant
on top of the rest of the \name{} pipeline on all three datasets. None of the three 
variants contributes positively on any dataset, except vanilla refinement on Neuropathic, which has an Edge \(F_1\)  delta of nearly zero.

\begin{table}[ht]
\centering
\scriptsize
\begin{tabular}{@{}lccc@{}}
\toprule
\textbf{Variant} & \textbf{Insurance} & \textbf{Hepar2} & \textbf{Neuropathic} \\
\midrule
Vanilla refinement & $-0.042$ & $-0.160$ & $0.001$ \\
Refinement + support & $-0.009$ & $-0.026$ & $-0.012$ \\
Narrowly scoped refinement & $-0.031$ & $-0.059$ & $-0.108$ \\
\bottomrule
\end{tabular}
\caption{\(F_1\) change when graph refinement is appended to the rest of
the \name{} pipeline, by variant and dataset.}
\label{tab:refinement-deltas}
\end{table}

For a representative example: on Hepar2 with the subsampling-support
variant, refinement proposed 21 deletes, of which 12 were
true-positive edges. Five of those true positives had subsampling
support $\ge 0.9$, including \texttt{bilirubin}$\to$\texttt{itching}
at support $0.991$. The prompt showed the LLM each edge's support
and explicitly framed high support as evidence the edge was unlikely
to be spurious, but the LLM proposed the deletes anyway. The narrow 
scope variant exhibits the same mechanism: of eight deletes proposed
on Hepar2 for confounder redirection, all eight were true-positive edges, 
and of eighteen proposed additional edges to connect orphan nodes, 
only one was true-positive.

The three variants in
Table~\ref{tab:refinement-deltas} span the natural axes of
remediation (prompt enrichment, action-space restriction, and a
combined justification requirement) and they converge to the same
failure mode. Once a high-precision graph has been assembled, every
remaining unilateral edit the LLM is empowered to make is fundamentally a calibration disagreement against statistical evidence the
LLM has already been shown. This implies that augmentations that compose
positively are the ones that contribute evidence (additional
sampling, sampling weights, parent endorsements that bypass
the cutoff). Augmentations that propose edits to an already-
assembled high-precision graph are likely to degrade results,
regardless of how they are scoped or informed.

\section{Stagewise Error Analysis}
\label{app:prompts}

\begin{table}[t]
\centering
\small
\resizebox{\linewidth}{!}{
\begin{tabular}{lccccc}
\toprule
\textbf{Stage} &
\textbf{Insurance} &
\textbf{Hepar2} &
\textbf{Neuropathic} &
\textbf{Diabetes} &
\textbf{Munin} \\
\midrule
\multicolumn{6}{l}{\textbf{ABSOL}} \\
\quad Candidate generation
    & $6.2\%$ & $6.5\%$ & $29.1\%$ & $0.7\%$ & $67.0\%$ \\
\quad Direction resolution
    & $4.2\%$ & $11.4\%$ & $7.2\%$ & $22.6\%$ & $8.1\%$ \\
\quad Parent selection
    & $1.9\%$ & $10.1\%$ & $26.5\%$ & $3.7\%$ & $0.0\%$ \\
\quad Greedy assembly
    & $0.0\%$ & $0.3\%$ & $0.3\%$ & $16.9\%$ & $2.6\%$ \\
\quad \emph{Recovered}
    & $87.7\%$ & $71.7\%$ & $36.9\%$ & $56.1\%$ & $22.3\%$ \\
\midrule
\multicolumn{6}{l}{\textbf{ABSOL\textsubscript{noLLM}}} \\
\quad Candidate generation
    & $6.2\%$ & $10.0\%$ & $33.4\%$ & $30.9\%$ & $87.1\%$ \\
\quad Vote threshold
    & $13.5\%$ & $22.2\%$ & $5.4\%$ & $1.2\%$ & $0.0\%$ \\
\quad Direction resolution
    & $5.8\%$ & $10.4\%$ & $10.8\%$ & $29.4\%$ & $2.5\%$ \\
\quad Greedy assembly
    & $0.0\%$ & $0.0\%$ & $27.6\%$ & $30.1\%$ & $7.5\%$ \\
\quad \emph{Recovered}
    & $74.5\%$ & $57.5\%$ & $22.8\%$ & $8.5\%$ & $2.9\%$ \\
\bottomrule
\end{tabular}
}
\caption{
Percentage of ground-truth edges that are lost at each stage, with Recovered as the final directed-edge recall. Only Candidate Generation and Recovered are directly comparable across methods because the intermediate stages operate on different surviving candidate sets. \textsc{Insurance}, \textsc{Hepar2}, and \textsc{Neuropathic} are averaged over five trials, while \textsc{Diabetes} and \textsc{Munin} are single runs. All \name numbers are with GPT-5.4, and are from the same runs as in Table~\ref{tab:main-results} and Table~\ref{tab:augmentation-ablation}.
}
\label{tab:module_diagnostics}
\end{table}

Table~\ref{tab:augmentation-ablation} measures each augmentation's end-to-end contribution through leave-one-out ablations. Because the four LLM interfaces perform different conditional tasks, raw per-call accuracy measures are conditioned on different candidate sets and are not directly comparable across augmentations. Instead, to localize the system-level errors to specific stages and identify process bottlenecks, we analyze the retained intermediate states from the existing runs. In Table~\ref{tab:module_diagnostics}, for each ground-truth edge, we assign its loss to the first stage after which it can no longer appear in the final graph.

The decomposition shows that there is no single downstream bottleneck: the dominant loss stage varies substantially across datasets and scales. The matched comparison shows where the final recall gains first become possible, while the stage decomposition localizes the points at which ground-truth edges become irrecoverable. The final recall of recovered ground-truth edges rises uniformly with LLM augmentations across the five benchmarks, with an average gain of $+0.217$. On the two largest graphs, much of this gain occurs before final assembly: candidate-generation losses fall from $30.9\%$ to $0.7\%$ on \textsc{Diabetes} and from $87.1\%$ to $67.0\%$ on \textsc{Munin}. This complements the leave-one-out results, 
where removing Parent Ordering lowers Edge \(F_1\) on all three repeated-run benchmarks, with the largest drop on \textsc{Insurance} and \textsc{Neuropathic} and the second-largest on \textsc{Hepar2}.

\section{Prompt Templates}
\label{app:prompts}

\begin{promptbox}{Shared preamble (prepended to every system message)}
You are an expert on causal reasoning with deep subject matter expertise. You are working on creating a new Bayesian Network.

Reason carefully about direct causal relationships between variables, distinguishing them from indirect effects, common-cause confounding, and coincidental correlation. When frequencies or candidate lists from statistical learning are provided, treat them as informative but not authoritative - they may reflect sampling variance or confounding variables.
\end{promptbox}

\subsection{Column Grouping}

\begin{promptbox}{System message}
We are building a Bayesian Network and need to identify which other variables are likely to have a direct causal relationship with a target variable. This will be used to bias sampling so that causally related variables are more often sampled together.

Identify the variables that are likely to be directly causally linked to the target variable - either as a direct cause of the target variable or as a direct effect of the target variable. Exclude variables whose relationship with the target variable is only indirect (mediated through other variables) or purely correlational without a direct causal mechanism.

Respond with only a Python list of variable names, nothing else. Example format:
["NodeA", "NodeB", "NodeC"]
\end{promptbox}

\begin{promptbox}{User message}
{variable_descriptions}

The target variable is: {target_node}

Reminder: respond with only a Python list of variable names, nothing else.
\end{promptbox}

\subsection{Adaptive Sampling}

\begin{promptbox}{System message}
We are refining a Bayesian Network through subsample aggregation. For one specific node, the aggregation analysis has produced ambiguous results about its parent set, and we want to run additional targeted  samples to resolve the ambiguity.

You will be shown a target node, the candidate parents whose status remains uncertain, and the subsampling frequency of each candidate (the fraction of samples - in which both nodes co-occurred - where the candidate appeared as a parent of the target node). Frequencies near 0.5 indicate the samples disagree on whether the edge belongs in the final graph.

Identify which OTHER columns should be co-sampled with the target node in additional samples to better resolve which of the uncertain candidate parents are TRUE direct causes of the target node. Focus on:
1. Likely confounders - variables that may be common causes of the target node and one or more of the uncertain candidates, whose presence would let the structure learner correctly attribute the relationship.
2. Likely mediators - variables sitting causally between the target node and a candidate, whose inclusion clarifies whether the candidate is a direct or indirect cause.
3. Disambiguators - variables causally linked to one candidate but not others, helping distinguish which candidates reflect true direct relationships.

Do NOT include the target node itself or the uncertain candidates already listed - those will already be included. Pick 3-8 columns from the available list. If you cannot identify useful co-sampling columns, return an empty list.

Respond with only a Python list of column names, nothing else. Example format:
["ColumnA", "ColumnB", "ColumnC"]
\end{promptbox}

\begin{promptbox}{User message}
{variable_descriptions}

TARGET NODE: {target_node}

The subsample aggregation analysis identified the following candidate parents whose status remains uncertain.

{uncertain_edges}

Reminder: respond with only a Python list of column names, nothing else.
\end{promptbox}

\subsection{Parent Ordering}

\begin{promptbox}{System message}
We need to determine which nodes should be the parents of a target node in our Bayesian Network, as well as the order of importance for these parents.

We have selected the following candidate parent nodes by training many small Bayesian Networks on subsets of the data.

Take these percentages into account, but do not blindly follow them. They could be easily swayed by random variance in sampling frequency - only a small number of nodes are present in each sample - or by spurious correlations in the data subsets.

Respond only with a Python-formatted list of your chosen parent node names, nothing else. The list should be in order of relevance, with the strongest parent candidates ranked first. Make sure the parent node names are exact string matches of the ones provided to you.
\end{promptbox}

\begin{promptbox}{User message}
{variable_descriptions}

Target Node: {target_node}

We are displaying the percentage of subsample networks in which the edge parent->{target_node} appeared.
{parent_percentages}

Reminder: respond only with a Python-formatted list of your chosen parent node names, nothing else.
\end{promptbox}

When adaptive sampling is also enabled, the system message above is
extended with the following block, placed before the response-format
instruction. The block explains the additional pre-/post-
adaptive-sampling trajectory annotations that the user message attaches
to any candidate the adaptive-sampling step targeted.

\begin{promptbox}{Additional system block (adaptive sampling enabled)}
Some candidates may be annotated with a pre-AS / post-AS trajectory of the form:
    <parent>: <pct> (<count>/<co_occurrence> networks where both co-occurred)
        pre-AS:  <pct> (<count>/<co_occurrence> initial subsample networks where both co-occurred) - AS triggered at H=<entropy>
        post-AS: <pct> (<count>/<co_occurrence> AS-targeted samples where both co-occurred)
These candidates were flagged by an upstream "adaptive sampling" (AS) step because the initial subsample network was uncertain about them (their pre-AS Bernoulli entropy H exceeded a threshold; H is in bits, 1.0 = maximum uncertainty / 50-50, 0.0 = unanimous). The pipeline then ran additional samples that forced the target node and the uncertain candidates to be co-sampled, with the intent of resolving the ambiguity. The "post-AS" line shows how the edge fared in those targeted samples in isolation, and the "overall" line is the combined evidence.

When a candidate has this trajectory, the targeted samples were drawn specifically to disambiguate it, so the post-AS rate is informative about whether the relationship survives controlled co-sampling. A candidate where post-AS rate is high and AS triggered at high H is one the pipeline deliberately scrutinized and where the targeted evidence supports the edge.
\end{promptbox}

\subsection{Cycle Arbitration}

\begin{promptbox}{System message}
We are assembling a Bayesian Network by greedily adding directed edges. When a proposed new edge would close one or more cycles, we must remove one or more existing edges so that NO directed path from the new edge's child back to its parent remains.

Pick the smallest set of edges to drop such that every offending path has at least one of its edges removed. Prefer dropping edges that are least likely to reflect a true direct causal relationship - edges with implausible direction or pairs more likely to be correlated without direct causation - over edges that reflect well-established domain knowledge.

If no acceptable removal set exists (i.e. you would rather keep all existing edges and not add the proposed new edge), respond with an empty list.

Respond with only a Python list of edge numbers, nothing else. Example: [2, 5]
\end{promptbox}

\begin{promptbox}{User message}
{variable_descriptions}

We are about to add the edge {new_edge_parent} -> {new_edge_child}, but this would create one or more cycles because the following directed paths from {new_edge_child} to {new_edge_parent} already exist in the graph:

{cycle_paths}

If we add {new_edge_parent} -> {new_edge_child}, every path above closes into a cycle.

The candidate edges to remove (the union of all edges appearing in any path above) are:
{edge_options}

Reminder: respond with only a Python list of edge numbers, nothing else.
\end{promptbox}

\section{Artifact License}
\label{app:license}

Code, prompts, and experiment configurations will be released under the BSD 3-Clause License.
Third-party benchmark networks, software libraries, and hosted LLM outputs remain subject to their original licenses and terms of use.

\end{document}